\documentclass[a4paper,fleqn]{cas-sc}

\usepackage[numbers,sort&compress]{natbib}
\usepackage{algorithm}
\usepackage{algorithmic}
\usepackage{array}
\usepackage{textcomp}
\usepackage{url}
\usepackage{verbatim}
\usepackage{amsmath}
\usepackage{amsthm}
\usepackage{amsfonts}
\usepackage{extarrows}
\usepackage{graphicx}
\usepackage{epsfig}
\usepackage{amssymb}
\usepackage{mathtools}
\usepackage{tabularx}
\usepackage{bm}
\usepackage{makecell}
\usepackage{hhline}
\usepackage{booktabs}
\usepackage{harpoon}
\usepackage{epstopdf}
\usepackage{cases}
\usepackage{threeparttable}
\usepackage{multicol}
\usepackage{multirow}

\hypersetup{hypertexnames=false}

\begin{document}
\let\WriteBookmarks\relax
\def\floatpagepagefraction{1}
\def\textpagefraction{.001}

\shorttitle{Rotation-Parameterized Graph Fractional Fourier Transform}
\shortauthors{F. Zhao et~al.}

\title[mode = title]{Rotation-Parameterized Graph Fractional Fourier Transform: Definition, Properties, and Optimal Filtering}
\tnotemark[1]

\tnotetext[1]{This work was supported in part by the Open Foundation of Hubei Key Laboratory of Applied Mathematics (Hubei University) under Grant HBAM202404; in part by the Foundation of Key Laboratory of System Control and Information Processing, Ministry of Education under Grant Scip20240121; and in part by the Startup Foundation for Introducing Talent of Nanjing Institute of Technology under Grant YKJ202214. \emph{(Corresponding author: Zhichao~Zhang.)}}

\author[1]{Feiyue~Zhao}
\ead{202511150010@nuist.edu.cn}

\author[1]{Mingzhi~Wang}
\ead{wmz200208@163.com}

\author[2,3,4]{Yangfan~He}
\ead{Yangfan.He@njit.edu.cn}

\author[1,5,6]{Zhichao~Zhang}
\cormark[1]
\ead{zzc910731@163.com}

\affiliation[1]{organization={School of Mathematics and Statistics, Nanjing University of Information Science and Technology},
            city={Nanjing},
            postcode={210044},
            country={China}}

\affiliation[2]{organization={School of Communication and Artificial Intelligence, Nanjing Institute of Technology},
            city={Nanjing},
            postcode={211167},
            country={China}}

\affiliation[3]{organization={School of Integrated Circuits, Nanjing Institute of Technology},
            city={Nanjing},
            postcode={211167},
            country={China}}

\affiliation[4]{organization={Jiangsu Province Engineering Research Center of IntelliSense Technology and System},
            city={Nanjing},
            postcode={211167},
            country={China}}

\affiliation[5]{organization={Hubei Key Laboratory of Applied Mathematics, Hubei University},
            city={Wuhan},
            postcode={430062},
            country={China}}

\affiliation[6]{organization={Key Laboratory of System Control and Information Processing, Ministry of Education, Shanghai Jiao Tong University},
            city={Shanghai},
            postcode={200240},
            country={China}}

\cortext[1]{Corresponding author}

\begin{abstract}
Graph spectral representations are fundamental in graph signal processing, providing a rigorous framework for analyzing graph-structured data. The graph fractional Fourier transform (GFRFT) extends the graph Fourier transform (GFT) through a fractional-order parameter, enabling flexible spectral analysis with mathematical consistency. The angular graph Fourier transform (AGFT) further introduces angular control by rotating GFT eigenvectors; however, existing constructions may fail to reduce exactly to the GFT at zero angle, weakening theoretical consistency and interpretability. To address these complementary limitations, namely the lack of rotation-based basis control in GFRFT and the defective zero-angle degeneracy of AGFT, this paper proposes the rotation-parameterized graph fractional Fourier transform (RP-GFRFT), which unifies fractional order and rotation-parameterized spectral analysis. A degeneracy-preserving rotation matrix family is constructed to guarantee exact GFT reduction at zero angle. Two RP-GFRFT variants, I-RP-GFRFT and II-RP-GFRFT, are then formulated, with theoretical analyses confirming their unitarity, invertibility, reduction behavior, and smooth parameter dependence. The fractional order and rotation angle are jointly optimized for adaptive graph spectral filtering. Experiments on real-world signals, images, and point clouds demonstrate that RP-GFRFT improves denoising accuracy, reconstruction quality, and feature preservation over GFRFT, AGFT, and representative filtering baselines.

\end{abstract}


\begin{NoHyper}
\maketitle
\end{NoHyper}

\section{Introduction}
Graph signal processing (GSP) has emerged as a foundational framework for analyzing non-Euclidean structured data~\cite{bg00, bg02new, bg03, bg01, bg04, bg05, bggnn0, bg06, bgnew0, bgnew1,tay2024kernelggsp,tay2025networkmht,tay2025distributionalgnn,routtenberg2021nonbayesian,routtenberg2022bayesian,bg19}, addressing the inherent limitations of traditional signal processing techniques that are tailored exclusively to regular grid signals (e.g., time series, images). Unlike classical methods, which rely on shift-invariance and uniform topology to enable rigorous spectral analysis, conventional signal processing tools cannot be directly extended to graph-structured data where signals are defined on irregular vertex sets with complex relational dependencies~\cite{bggr01, bggr02, bggr04}. This critical gap has driven the rapid advancement of GSP, which generalizes core concepts of classical signal processing (e.g., Fourier transform, filtering, spectral decomposition) to the graph domain, thereby providing principled and interpretable tools for transforming, analyzing, and enhancing signals supported on graphs~\cite{spGraphHilbert2019,spGraphSketching2020}. As GSP gains growing utility in high-impact applications~\cite{bg16, bg13, bg14, bg17, bg19, bg20}, including image processing, spatiotemporal sensor data analysis, and 3D point cloud denoising, the demand for more flexible and theoretically consistent spectral transformation methods has become increasingly pressing.

At the core of GSP lies the graph spectral representation, where graph signals are projected onto the eigenspace of a graph shift operator (e.g., an adjacency matrix)~\cite{bg07, bg08, bg09, bg10, bg11, bg12}. The resulting graph Fourier transform (GFT) reveals the intrinsic frequency components of the signals, thereby enabling efficient filtering, denoising, compression, and spectral analysis of graph-structured data~\cite{gft0, gft1, gft2, gft3, gft4, gft5, gft6, gft7}. The proven effectiveness of GFT across diverse graph signal processing tasks underscores the pivotal role of spectral representations in rigorous graph-based signal analysis. However, as the scale and complexity of graph-structured data increase, the limitations of conventional GFT become increasingly prominent, driving the need for spectral transforms with enhanced flexibility, adaptability, and interpretability that align with the demands of complex graph signal processing scenarios.

To address this demand, fractional spectral methods have been explored, with the graph fractional Fourier transform (GFRFT) extending the classical GFT by introducing a fractional-order parameter~\cite{gfrft0, gfrft1, gfrft2, gfrft3, gfrft8, gfrft4, gfrft5, gfrft6}. This parameter generates intermediate spectral domains that smoothly interpolate between the vertex and graph-frequency domains, providing an additional degree of freedom to enable finer spectral resolution and adaptive filtering. Recent joint time-vertex extensions have further shown that fractional Fourier analysis can be combined with graph spectral tools to process time-varying graph signals~\cite{gfrft7}. These capabilities surpass the analytical limits of conventional GFT while preserving mathematical rigor. However, GFRFT lacks mechanisms for rotation-based manipulation of the spectral basis, which is a critical limitation in applications requiring continuous control over the spectral orientation. To mitigate this limitation, the angular graph Fourier transform (AGFT) was proposed to incorporate angular control via the rotation of GFT eigenvectors based on prescribed rotation parameters, adding another dimension of flexibility for spectral representation and controllable transformation domains~\cite{agft}. Nevertheless, existing AGFT constructions typically fail to degenerate to the original GFT at zero angle, undermining both theoretical consistency and interpretability, and are not inherently integrated with fractional-order analysis, limiting adaptability in scenarios where both rotation and fractional flexibility are essential.
\begin{figure}[pos=t]
\centering
\includegraphics[width=\textwidth]{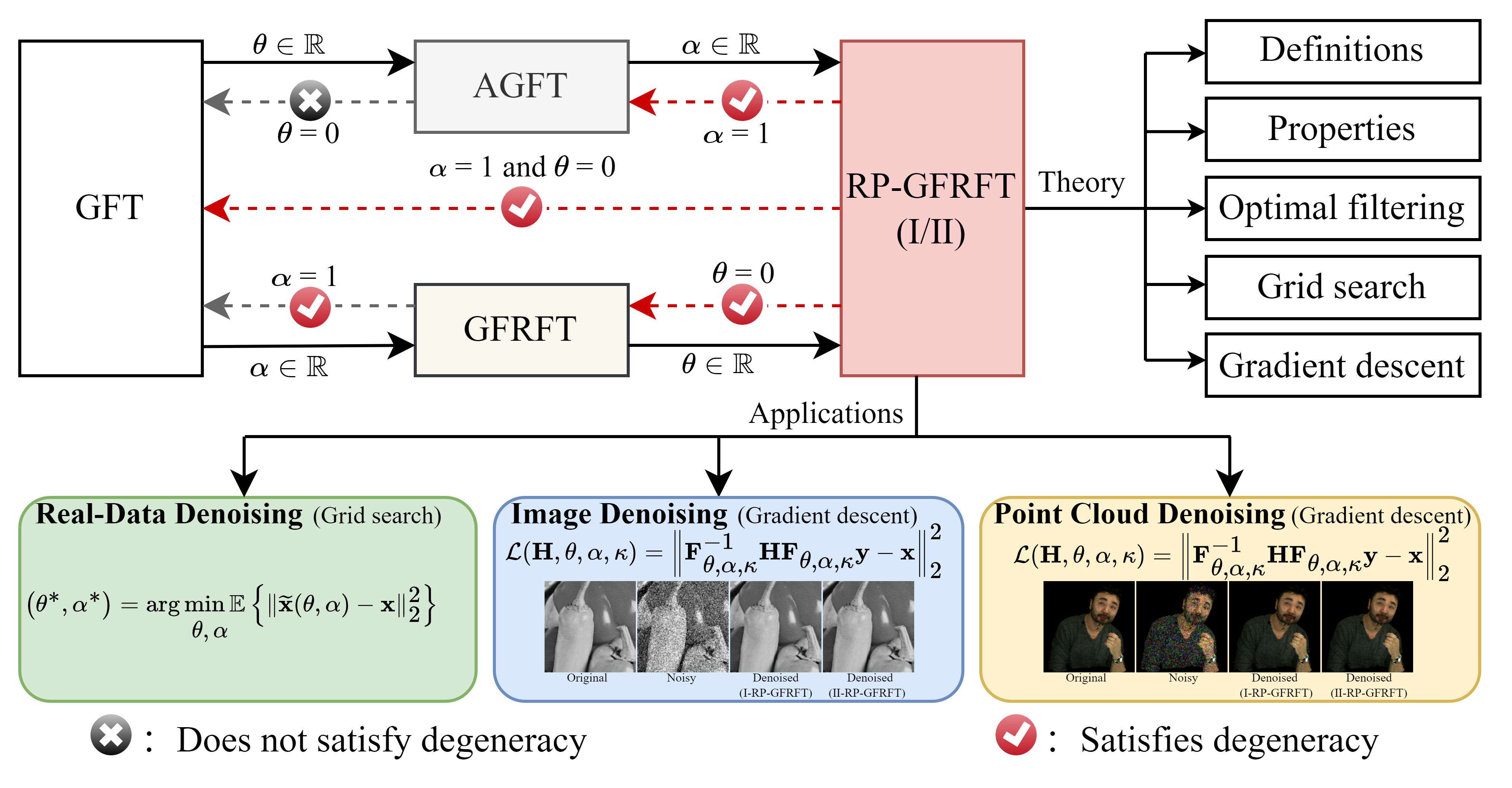}
\caption{Overview of the proposed RP-GFRFT framework and its applications.}
\label{fig:framework}
\end{figure}

To overcome these complementary limitations of GFRFT and AGFT, this paper introduces the rotation-parameterized graph fractional Fourier transform (RP-GFRFT), a unified framework that integrates fractional-order and rotation-parameterized spectral analysis while maintaining theoretical consistency and practical implementability. The overall architecture of the proposed RP-GFRFT framework is illustrated in Figure~\ref{fig:framework}, with its key contributions summarized as follows:
\begin{itemize}
	\item{We design a degeneracy-friendly rotation matrix family ensuring exact GFT reduction at zero angle, resolving conventional AGFT’s consistency and interpretability flaws.}
	\item{We introduce two RP-GFRFT formulations (I-RP-GFRFT/II-RP-GFRFT) and rigorously analyze their core properties and smooth variation with both rotation and fractional-order parameters.}
	\item{We provide a learnable rotation-fractional order joint parametrization, enabling optimized spectral processing via standard gradient-based methods.}
	\item{We validate the framework via extensive experiments, demonstrating superiority over GFRFT and AGFT in spectral concentration, reconstruction quality, and controllable manipulation.}
\end{itemize}

The remainder of this paper is organized as follows: Section \uppercase\expandafter{\romannumeral 2} reviews preliminaries. Section \uppercase\expandafter{\romannumeral 3} details the proposed RP-GFRFT framework. Section \uppercase\expandafter{\romannumeral 4} presents the experimental results and analysis. Finally, Section \uppercase\expandafter{\romannumeral 5} concludes the paper and outlines future research directions.

\section{Preliminaries}
This section introduces the core GSP concepts that form the foundation of the proposed RP-GFRFT. We first review graph signals and shift operators, then cover GFT and GFRFT, and conclude with basis rotations in the graph spectral space and AGFT.
\subsection{Graph Signals and Graph Shift Operators}

Let $\mathcal{G}=(\mathcal{V}, \mathcal{E}, \mathbf{W})$ denote an undirected weighted graph, where $\mathcal{V}$ is the set of $N$ nodes, $\mathcal{E}$ is the edge set, and $\mathbf{W} \in \mathbb{R}^{N\times N}$ is the weighted adjacency matrix. A graph signal $\mathbf{x} \in \mathbb{R}^N$ assigns a scalar value to each node of the graph~\cite{bg03, bg13}.

Graph shift operators (GSOs), such as the adjacency matrix $\mathbf{A}$, weighted adjacency $\mathbf{W}$, or graph Laplacian $\mathbf{L}$, encode the underlying graph structure and provide the foundation for spectral analysis~\cite{bg02}.

\subsection{Graph Fourier Transform (GFT)}

Consider the eigendecomposition of a general GSO $\mathbf{Z}$:
\begin{equation}
\mathbf{Z} = \mathbf{U} \mathbf{\Lambda} \mathbf{U}^{-1} = \mathbf{U} \mathbf{\Lambda} \mathbf{U}^{\mathsf{H}},
\label{eq1}
\end{equation}
where $\mathbf{\Lambda} = \mathrm{diag}(\lambda_1, \dots, \lambda_N)$ contains the eigenvalues and $\mathbf{U} = [\mathbf{u}_1, \dots, \mathbf{u}_N]$ contains the corresponding eigenvectors. The graph Fourier transform of a signal $\mathbf{x}$ is defined as
\begin{equation}
\widehat{\mathbf{x}} = \mathbf{F} \mathbf{x}, \quad \mathbf{F} = \mathbf{U}^{-1} = \mathbf{U}^{\mathsf{H}}.
\label{eq2}
\end{equation}
The eigenvectors $\mathbf{u}_k$ form an orthonormal basis spanning the graph spectral domain, enabling filtering, analysis, and representation of graph signals~\cite{gft0, gft3}.

\subsection{Graph Fractional Fourier Transform (GFRFT)}

The GFRFT introduces a fractional-order parameter $\alpha \in \mathbb{R}$, which interpolates between the vertex and graph-frequency domains:
\begin{equation}
\widehat{\mathbf{x}}^\alpha = \mathbf{F}^\alpha \mathbf{x}.
\label{eq3}
\end{equation}
By varying $\alpha$, one can flexibly control the spectral repmresentation while remaining mathematically consistent with the GFT~\cite{gfrft0, gfrft3}.

\subsection{Roll, Pitch, and Yaw Rotation Matrices}
Assume that $\mathbf{R}_{M\times M}$ is an $M\times M$ orthogonal matrix, then the matrices
\begin{equation}
\mathbf{R}_{2M\times 2M}
=
\frac{1}{\sqrt{2}}
\begin{pmatrix}
\mathbf{R}_{M\times M} & \mathbf{R}_{M\times M} \\
-\mathbf{R}^{\cap}_{M\times M} & \mathbf{R}^{\cap}_{M\times M}
\end{pmatrix},
\label{eq4} 
\end{equation}
\begin{equation}
\mathbf{R}^{\mathrm{roll}}_{(2M+1)\times(2M+1)} =
\frac{1}{\sqrt{2}}
\begin{pmatrix}
\sqrt{2} & \mathbf{0}_{1\times M} & \mathbf{0}_{1\times M} \\
\mathbf{0}_{M\times 1} & \mathbf{R}_{M\times M} & \mathbf{R}_{M\times M} \\
\mathbf{0}_{M\times 1} & -\mathbf{R}^{\cap}_{M\times M} & \mathbf{R}^{\cap}_{M\times M}
\end{pmatrix},
\label{eq5}
\end{equation}
\begin{equation}
\mathbf{R}^{\mathrm{pitch}}_{(2M+1)\times(2M+1)} =
\frac{1}{\sqrt{2}}
\begin{pmatrix}
\mathbf{R}_{M\times M} & \mathbf{0}_{M\times 1} & -\mathbf{R}^{\cap}_{M\times M} \\
\mathbf{0}_{1\times M} & \sqrt{2} & \mathbf{0}_{1\times M} \\
\mathbf{R}_{M\times M} & \mathbf{0}_{M\times 1} & \mathbf{R}^{\cap}_{M\times M}
\end{pmatrix},
\label{eq6}
\end{equation}
\begin{equation}
\mathbf{R}^{\mathrm{yaw}}_{(2M+1)\times(2M+1)} =
\frac{1}{\sqrt{2}}
\begin{pmatrix}
\mathbf{R}_{M\times M} & \mathbf{R}_{M\times M} & \mathbf{0}_{M\times 1} \\
- \mathbf{R}^{\cap}_{M\times M} & \mathbf{R}^{\cap}_{M\times M} & \mathbf{0}_{M\times 1} \\
\mathbf{0}_{1\times M} & \mathbf{0}_{1\times M} & \sqrt{2}
\end{pmatrix}.
\label{eq7}
\end{equation}
are both orthogonal matrices, where $\mathbf{0}_{M\times1}$ and $\mathbf{0}_{1\times M}$ are $M\times1$ and $1\times M$ zero vectors, respectively, and 
\(
\mathbf{R}^{\cap}_{M\times M} := \big[r_{\,M+1-i,\,j}\big]_{i,j=1,2,\cdots,M}
\)
is defined by upside down the matrix
\(
\mathbf{R}_{M\times M} = \big[r_{\,i,\,j}\big]_{i,j=1,2,\cdots,M}.
\)

The \(2\times2\) rotation matrix reads
\begin{equation}
\mathbf{R}^{\theta}_{2\times2} =
\begin{pmatrix}
\cos\theta & \sin\theta \\
-\sin\theta & \cos\theta
\end{pmatrix},
\label{eq8}
\end{equation}
where \(\theta\in[0,2\pi]\). The following three basic rotation matrices that rotate vectors by a roll, pitch, or yaw angle \(\theta\) about the $x$-, $y$-, or $z$-axis, in three dimensions:
\begin{equation}
\mathbf{R}_{3\times3}^{\mathrm{roll},\theta} =
\begin{pmatrix}
1 & 0 & 0 \\
0 & \cos\theta & \sin\theta \\
0 & -\sin\theta & \cos\theta
\end{pmatrix},
\label{eq9}
\end{equation}

\begin{equation}
\mathbf{R}_{3\times3}^{\mathrm{pitch},\theta} =
\begin{pmatrix}
\cos\theta & 0 & -\sin\theta \\
0 & 1 & 0 \\
\sin\theta & 0 & \cos\theta
\end{pmatrix},
\label{eq10}
\end{equation}

\begin{equation}
\mathbf{R}_{3\times3}^{\mathrm{yaw},\theta} =
\begin{pmatrix}
\cos\theta & \sin\theta & 0 \\
-\sin\theta & \cos\theta & 0 \\
0 & 0 & 1
\end{pmatrix},
\label{eq11}
\end{equation}
where \(\theta\in[0,2\pi]\)~\cite{rm0,rm1,rm2,rm3}. 

Due to Eqs. (\ref{eq4}), (\ref{eq5}), (\ref{eq8}) and (\ref{eq9}), the \(N \times N\) rotation matrix \(\mathbf{R}_{N \times N}^{\mathrm{roll}, \theta}\) that rotates vectors by a roll angle \(\theta\) about the \(x\)-axis, abbreviated as the \(N \times N\) roll rotation matrix, can be computed iteratively as
\begin{equation}
\mathbf{R}_{2M\times 2M}^{\mathrm{roll},\theta} =
\frac{1}{\sqrt{2}}
\begin{pmatrix}
\mathbf{R}_{M\times M}^{\mathrm{roll},\theta} & \mathbf{R}_{M\times M}^{\mathrm{roll},\theta} \\
-(\mathbf{R}_{M\times M}^{\mathrm{roll},\theta})^{\cap} & (\mathbf{R}_{M\times M}^{\mathrm{roll},\theta})^{\cap}
\end{pmatrix}
\label{eq12}
\end{equation}
and
\begin{equation}
\begin{aligned}
&\mathbf{R}_{(2M+1)\times(2M+1)}^{\mathrm{roll},\theta}=
\frac{1}{\sqrt{2}}
\begin{pmatrix}
\sqrt{2} & \mathbf{0}_{1\times M} & \mathbf{0}_{1\times M} \\
\mathbf{0}_{M\times 1} & \mathbf{R}_{M\times M}^{\mathrm{roll},\theta} & \mathbf{R}_{M\times M}^{\mathrm{roll},\theta} \\
\mathbf{0}_{M\times 1} & -(\mathbf{R}_{M\times M}^{\mathrm{roll},\theta})^{\cap} & (\mathbf{R}_{M\times M}^{\mathrm{roll},\theta})^{\cap}
\end{pmatrix}
\label{eq13}
\end{aligned}
\end{equation}
for even \(N = 2M\) and odd \(N = 2M+1\), respectively. Similarly, the \(N \times N\) pitch and yaw rotation matrices \(\mathbf{R}_{N \times N}^{\mathrm{pitch}, \theta}\) and \(\mathbf{R}_{N \times N}^{\mathrm{yaw}, \theta}\) can be computed iteratively as 
\begin{equation}
\mathbf{R}_{2M\times 2M}^{\mathrm{pitch},\theta} =
\frac{1}{\sqrt{2}}
\begin{pmatrix}
\mathbf{R}_{M\times M}^{\mathrm{pitch},\theta} & \mathbf{R}_{M\times M}^{\mathrm{pitch},\theta} \\
-(\mathbf{R}_{M\times M}^{\mathrm{pitch},\theta})^{\cap} & (\mathbf{R}_{M\times M}^{\mathrm{pitch},\theta})^{\cap}
\end{pmatrix},
\label{eq14}
\end{equation}
\begin{equation}
\begin{aligned}
&\mathbf{R}_{(2M+1)\times(2M+1)}^{\mathrm{pitch},\theta}=
\frac{1}{\sqrt{2}}
\begin{pmatrix}
\mathbf{R}_{M\times M}^{\mathrm{pitch},\theta} & \mathbf{0}_{M\times 1} & -(\mathbf{R}_{M\times M}^{\mathrm{pitch},\theta})^{\cap} \\
\mathbf{0}_{1\times M} & \sqrt{2} & \mathbf{0}_{1\times M} \\
\mathbf{R}_{M\times M}^{\mathrm{pitch},\theta} & \mathbf{0}_{M\times 1} & (\mathbf{R}_{M\times M}^{\mathrm{pitch},\theta})^{\cap}
\end{pmatrix}
\label{eq15}
\end{aligned}
\end{equation}
and
\begin{equation}
\mathbf{R}_{2M\times 2M}^{\mathrm{yaw},\theta} =
\frac{1}{\sqrt{2}}
\begin{pmatrix}
\mathbf{R}_{M\times M}^{\mathrm{yaw},\theta} & \mathbf{R}_{M\times M}^{\mathrm{yaw},\theta} \\
-(\mathbf{R}_{M\times M}^{\mathrm{yaw},\theta})^{\cap} & (\mathbf{R}_{M\times M}^{\mathrm{yaw},\theta})^{\cap}
\end{pmatrix},
\label{eq16}
\end{equation}
\begin{equation}
\begin{aligned}
&\mathbf{R}_{(2M+1)\times(2M+1)}^{\mathrm{yaw},\theta}=
\frac{1}{\sqrt{2}}
\begin{pmatrix}
\mathbf{R}_{M\times M}^{\mathrm{yaw},\theta} & \mathbf{R}_{M\times M}^{\mathrm{yaw},\theta} & \mathbf{0}_{M\times 1} \\
-(\mathbf{R}_{M\times M}^{\mathrm{yaw},\theta})^{\cap} & (\mathbf{R}_{M\times M}^{\mathrm{yaw},\theta})^{\cap} & \mathbf{0}_{M\times 1} \\
\mathbf{0}_{1\times M} & \mathbf{0}_{1\times M} & \sqrt{2}
\end{pmatrix},
\label{eq17}
\end{aligned}
\end{equation}
respectively. Here, \(\mathbf{R}_{2M \times 2M}^{\text{roll}, \theta} = \mathbf{R}_{2M \times 2M}^{\text{pitch}, \theta} = \mathbf{R}_{2M \times 2M}^{\text{yaw}, \theta} = \mathbf{R}_{2M \times 2M}^{\theta}\).

Notably, the prototype matrices (Eqs. (\ref{eq8})–(\ref{eq11})), specifically designed for \(N=3\), naturally degenerate to the identity matrix \(\mathbf{I}_3\) at \(\theta=0\) (direct substitution of \(\cos0=1\) and \(\sin0=0\) yields \(\mathbf{I}_3\)). However, this critical degeneracy property is not preserved by the recursive high-dimensional rotation matrices (Eqs. (\ref{eq12})–(\ref{eq17})) when \(N \geq 4\).

\subsection{Angular Graph Fourier Transform (AGFT)}
Let $\mathbf{R}^{\rm axis}_N(\theta)$ be any orthogonal rotation matrix generated by one of the families above.  
Define the rotated eigenbasis
\begin{equation}
\mathbf{U}_{\theta}=\mathbf{R}^{\rm axis}_N(\theta)\mathbf{U},\qquad 
\mathbf{U}_{\theta}^{\!H}\mathbf{U}_{\theta}=\mathbf{I}.
\label{eq18}
\end{equation}
The angular graph Fourier transform (AGFT) of a graph signal \(\mathbf{x}\) associated with angle \(\theta\in[0,2\pi]\) is
\begin{equation}
\widehat{\mathbf{x}}_{\theta}=\mathbf{F}_{\theta}\mathbf{x},
\qquad 
\mathbf{F}_{\theta}=\mathbf{U}_{\theta}^{\!H}.
\label{eq19}
\end{equation}
AGFT retains unitarity for any fixed \(\theta\)~\cite{agft}. However, when high-dimensional rotation matrices \(\mathbf{R}^{\rm axis}_N(\theta)\) is taken from the recursive families above, one generally has \(\mathbf{F}_{\theta=0}\neq \mathbf{U}^{\!H}\), i.e., AGFT does not revert to the original GFT at zero angle. This lack of degeneracy undermines interpretability and motivates the degeneracy-friendly construction developed in the next section.

\section{Rotation-Parameterized Graph Fractional Fourier Transform (RP-GFRFT)}
\label{sec:rpgfrft}
This section elaborates on the proposed RP-GFRFT. We first motivate integrating rotation and fractional degrees of freedom to address limitations of existing graph spectral transforms, then introduce a degeneracy-friendly rotation matrix family ensuring exact GFT degeneration, formulate two concrete RP-GFRFT variants (Type I and Type II), establish their fundamental theoretical properties, and finally derive explicit, computationally tractable expressions for the derivatives required to learn the rotation and fractional parameters.
\begin{table*}[t]
\centering
\caption{Comparison of different graph spectral transforms.}
\renewcommand{\arraystretch}{1.2}
\resizebox{\linewidth}{!}{%
    \begin{tabular}{@{\hspace{2pt}}lcccccc@{\hspace{2pt}}}
        \hline
        \textbf{Method} & \textbf{Fractional order} & \textbf{Angular control} & \textbf{Degeneracy to GFT} & \textbf{Periodicity} & \textbf{Invertibility} & \textbf{Remarks} \\
        \hline
        GFRFT~\cite{gfrft0} & Yes & No & Yes & No & Yes & Spectral function extension of GFT \\
        AGFT~\cite{agft} & No & Yes & No & No & Yes & Rotation-based, inconsistent at zero angle \\
        Corrected AGFT (proposed) & No & Yes & Yes & Yes & Yes & Degeneracy-friendly rotation matrices \\
        RP-GFRFT (proposed) & Yes & Yes & Yes & Unknown & Yes & Unified framework (Type I/II) \\
        \hline
    \end{tabular}
}
\label{tab:spectral_transforms_comparison}
\end{table*}

\subsection{Motivation and Concept}
The GFT provides a fixed orthonormal spectral basis determined by a GSO. The GFRFT extends GFT by a fractional-order parameter $\alpha$, enabling intermediate spectral domains and more flexible spectral analysis. Independently, AGFT rotates the GFT eigenbasis by a rotation parameter $\theta$ to provide additional spectral degrees of freedom. However, many commonly used AGFT rotation constructions employ angle-independent equal-weight block combinations (e.g. $\tfrac{1}{\sqrt{2}}\begin{pmatrix}\mathbf{I}&\mathbf{I}\\-\mathbf{I}&\mathbf{I}\end{pmatrix}$) in their recursion; as a result they fail to degenerate to the original GFT at $\theta=0$ (a structural ``no-go''). This undermines interpretability and consistency: a transform that purports to be a rotation-based generalization should recover GFT when the angle vanishes.

Beyond overcoming the limitations of conventional rotation-based graph transforms in distinguishing high-dimensional rotations, a principal design objective of the RP-GFRFT is to incorporate adaptive parameter tuning for practical applications. Unlike fixed-transform methods, the parameters \(\theta\) and \(\alpha\) are conceived as inherently tunable. This design enables the RP-GFRFT to function as a generalized, parameterized framework, where these parameters can be adjusted to optimize performance for specific signals or tasks, thereby facilitating highly tailored spectral processing.

Our objective is to unify fractional and rotation control while preserving the following desiderata:
\begin{itemize}
  \item \textbf{Degeneration:} at $\theta=0$ the rotation mechanism must reduce to the identity so that RP-GFRFT$\big|_{\theta=0}=\,$GFRFT.
  \item \textbf{Group structure:} each rotation $\mathbf{R}^{\rm axis}_N(\theta)$ should lie in $SO(N)$ (orthogonal with determinant $+1$) for every $\theta$.
  \item \textbf{Differentiability:} $\theta\mapsto\mathbf{R}^{\rm axis}_N(\theta)$ and the resulting RP-GFRFT operators should be sufficiently smooth to admit gradient-based optimization of $\theta$ and $\alpha$.
\end{itemize}

Table~\ref{tab:spectral_transforms_comparison} concisely summarizes the key attributes of representative graph spectral transforms, clearly illustrating how the proposed RP-GFRFT unifies fractional-order flexibility and rotation control while resolving the degeneracy inconsistency of conventional AGFT and the lack of rotation-based manipulation in GFRFT. 

\subsection{Degeneracy-Friendly Rotation Matrix Family}

We introduce a recursively constructed family of rotation matrices that is
\emph{degeneracy-friendly} (reduces to identity at zero rotation) and
distinguishes three axes (yaw, pitch, roll) for $N=2^k$ dimensions.

\emph{Notation:}
Let $\mathbf{P}_M \in \mathbb{R}^{M\times M}$ denote the anti-diagonal permutation
matrix (ones on the anti-diagonal). For any $\mathbf{X} \in \mathbb{R}^{M\times M}$,
define its \emph{diamond} (double-flip) by
\[
\mathbf{X}^\diamond := \mathbf{P}_M\, \mathbf{X}\, \mathbf{P}_M.
\]
Note that $\mathbf{P}_M^\top = \mathbf{P}_M = \mathbf{P}_M^{-1}$ and if
$\mathbf{X} \in O(M)$ then $\mathbf{X}^\diamond \in O(M)$ with
$\det(\mathbf{X}^\diamond) = \det(\mathbf{X})$.

\emph{Block rotation primitives:}
For integer $M\ge1$ and angle $\phi\in\mathbb{R}$ define the even-dimensional
block rotation (a direct sum of two-dimensional Givens blocks)
\[
\mathbf{S}_M(\phi)
:=\begin{pmatrix}
\cos\phi\,\mathbf{I}_M & \sin\phi\,\mathbf{I}_M\\[4pt]
-\sin\phi\,\mathbf{I}_M & \cos\phi\,\mathbf{I}_M
\end{pmatrix}\in SO(2M),
\]
and for the odd-dimensional case define three block rotations acting on the
block layout $(M,1,M)$:
\[
\mathbf{T}_M^{(1,2)}(\phi)
:=\begin{pmatrix}
\cos\phi\,\mathbf{I}_M & \sin\phi\,\mathbf{I}_M & \mathbf{0}\\[4pt]
-\sin\phi\,\mathbf{I}_M & \cos\phi\,\mathbf{I}_M & \mathbf{0}\\[4pt]
\mathbf{0} & \mathbf{0} & 1
\end{pmatrix}\in SO(2M+1),
\]
and similarly $\mathbf{T}_M^{(1,3)}(\phi),\ \mathbf{T}_M^{(2,3)}(\phi)$ for the
other two block-pair rotations. Each of these satisfies
$\mathbf{S}_M(0)=\mathbf{I}_{2M}$ and $\mathbf{T}_M^{(\cdot)}(0)=\mathbf{I}_{2M+1}$.

\emph{Axis-dependent exponential map for high-dimensional distinction:}  
To ensure that the yaw, pitch, and roll rotation families remain distinct in $N=2^k$ dimensions , we introduce axis-dependent skew-symmetric matrices $\mathbf{J}_{\rm roll}, \mathbf{J}_{\rm yaw}, \mathbf{J}_{\rm pitch} \in \mathbb{R}^{N \times N}$ (which satisfies \(\mathbf{J}^\top = -\mathbf{J}\)) with different sparsity patterns. These matrices are defined as
\[
\begin{aligned}
(\mathbf{J}_{\rm roll})_{i,j} &= 
\begin{cases}
-1, & j = i+1\\
1, & i = j+1\\
0, & \text{otherwise}
\end{cases}, \quad i,j=1,\dots,N-1,
\end{aligned}
\]
\[
\begin{aligned}
(\mathbf{J}_{\rm yaw})_{i,j} &= 
\begin{cases}
-1, & j = i+1,\ i\ \text{odd}\\
1, & i = j+1,\ i\ \text{odd},\\
0, & \text{otherwise}
\end{cases}
\end{aligned}
\]
\[
\begin{aligned}
(\mathbf{J}_{\rm pitch})_{i,j} &= 
\begin{cases}
-1, & j = i+1,\ i\ \text{even}\\
1, & i = j+1,\ i\ \text{even}.\\
0, & \text{otherwise}
\end{cases}
\end{aligned}
\]
The corresponding rotation in high dimensions is obtained by applying the matrix exponential as a small perturbation after the block-diagonal rotations:
\begin{equation}
\mathbf{R} = \mathbf{R}_{\rm block} \, \exp(\phi \mathbf{J}),
\label{eq20}
\end{equation}
where angle $\phi\in\mathbb{R}$ controls the magnitude of the perturbation.
This design ensures three properties:  
\begin{itemize}
    \item \textbf{Axis separation:} Different sparsity patterns prevent yaw, pitch, and roll families from collapsing onto each other in higher dimensions.
    \item \textbf{Orthogonality preservation:} As $\mathbf{J}$ is skew-symmetric, $\exp(\phi \mathbf{J})$ is guaranteed to be orthogonal.

    \emph{Proof:} Let \(\mathbf{J}\) be a skew-symmetric matrix and \(\phi\) an arbitrary scalar. A fundamental property of the matrix exponential states that the transpose of the exponential is the exponential of the transposed matrix, giving
	\(
	\left(\exp(\phi \mathbf{J})\right)^\top = \exp\left((\phi \mathbf{J})^\top\right).
	\)
	Substituting the skew-symmetry of \(\mathbf{J}\) (\(\mathbf{J}^\top = -\mathbf{J}\)), we obtain
	\(
	\left(\exp(\phi \mathbf{J})\right)^\top = \exp(-\phi \mathbf{J}).
	\)
	Since \(\phi \mathbf{J}\) and \(-\phi \mathbf{J}\) commute (i.e., \((\phi \mathbf{J})(-\phi \mathbf{J}) = (-\phi \mathbf{J})(\phi \mathbf{J})\)), the product of their exponentials satisfies
	\(
	\exp(-\phi \mathbf{J}) \exp(\phi \mathbf{J}) = \exp(-\phi \mathbf{J} + \phi \mathbf{J}) = \exp(\mathbf{0}) = \mathbf{I},
	\)
	where \(\mathbf{I}\) is the identity matrix. This confirms \(\exp(\phi \mathbf{J})^\top \exp(\phi \mathbf{J}) = \mathbf{I}\), the defining condition for an orthogonal matrix. Thus, \(\exp(\phi \mathbf{J})\) is orthogonal. \qed
    \item \textbf{Physical Interpretability:} The sparsity patterns of the skew-symmetric matrices $\mathbf{J}_{\mathrm{roll}}$, $\mathbf{J}_{\mathrm{pitch}}$, and $\mathbf{J}_{\mathrm{yaw}}$ directly reflect the adjacent coupling inherent in physical rotations. In 3D, each elemental rotation (roll, pitch, yaw) acts only on two adjacent axes. This adjacency principle is preserved in our construction: each $\mathbf{J}$-matrix couples only specific pairs of adjacent dimensions (e.g., odd or even indices), maintaining a clear distinction between rotation types. The pattern naturally extends to higher dimensions ($N = 2^k$), ensuring that high-dimensional rotations remain physically interpretable as compositions of intuitive, low-dimensional transformations.
\end{itemize}

In this way, the exponential map serves as a controlled perturbation that maintains the geometric and physical meaning of the rotations while distinguishing axes in high-dimensional spaces.

\emph{Recursive construction:}
Let $\phi:\mathbb{R} \to \mathbb{R}$ be $C^1$ with $\phi(0) = 0$, e.g., $\phi(\theta) = \kappa \theta$.

\begin{itemize}
  \item Base cases:
\begin{align*}
\mathbf{R}_1(\theta) &= 1, \\
\mathbf{R}_2(\theta) &=
\begin{pmatrix}
\cos\theta & \sin\theta\\ 
-\sin\theta & \cos\theta
\end{pmatrix}, \\
\mathbf{R}_3^{\text{yaw}}(\theta) &=
\begin{pmatrix}
\cos\theta & \sin\theta & 0\\ 
-\sin\theta & \cos\theta & 0\\ 
0 & 0 & 1
\end{pmatrix}.
\end{align*}
Similarly define base 3D blocks for pitch and roll.
    
  \item If $N=2M$ is even:
    \begin{align*}
    \mathbf{R}^{\rm axis}_{2M}(\theta) = &
    \operatorname{blkdiag}\big(\mathbf{R}_M(\theta), \mathbf{R}_M^\diamond(\theta)\big)
    \times \mathbf{S}_M(\phi(\theta)) \times \exp(\phi(\theta)\mathbf{J}_{\rm axis}(2M)),
    \end{align*}
    where ${\rm axis} \in \{\rm yaw, pitch, roll\}$ and $\mathbf{J}_{\rm axis}$ ensures axis distinction.
    
  \item If $N=2M+1$ is odd, define axis-dependent variants:
    \[
    \begin{aligned}
    \mathbf{R}^{\rm yaw}_{2M+1}(\theta) &= 
        \operatorname{blkdiag}(\mathbf{R}_M, \mathbf{R}_M^\diamond, 1)\, \mathbf{T}_M^{(1,2)}(\phi),\\
    \mathbf{R}^{\rm pitch}_{2M+1}(\theta) &= 
        \operatorname{blkdiag}(\mathbf{R}_M, 1, \mathbf{R}_M^\diamond)\, \mathbf{T}_M^{(1,3)}(\phi),\\
    \mathbf{R}^{\rm roll}_{2M+1}(\theta) &= 
        \operatorname{blkdiag}(1, \mathbf{R}_M, \mathbf{R}_M^\diamond)\, \mathbf{T}_M^{(2,3)}(\phi).
    \end{aligned}
    \]
\end{itemize}

\emph{Theorem 1:}
Let $\phi \in C^1$ with $\phi(0) = 0$. Then $\mathbf{R}^{\rm axis}_N(\theta)$
satisfies:
\begin{enumerate}
  \item $\mathbf{R}^{\rm axis}_N(\theta) \in SO(N)$;
  \item $\mathbf{R}^{\rm axis}_N(0) = \mathbf{I}_N$;
  \item $\theta \mapsto \mathbf{R}^{\rm axis}_N(\theta)$ is $C^1$;
  \item For $N = 2^k$, the yaw/pitch/roll families remain distinct due to
        axis-dependent skew-symmetric exponential maps.
\end{enumerate}
\emph{Proof:}
Induction on $N$. Base cases $N=1,2,3$ are immediate. Assume the theorem holds for $M$, i.e., $\mathbf{R}_M, \mathbf{R}_M^\diamond \in SO(M)$, block rotations belong to $SO(2M)$ (or $SO(2M+1)$), and $\exp(\phi \mathbf{J}_{\rm axis}) \in SO(2M)$. For the inductive step, products of matrices in $SO(\cdot)$ preserve orthogonality and determinant 1, hence remain in $SO(N)$. At $\theta=0$, $\mathbf{R}^{\rm axis}_N(0)$ reduces to the identity matrix by construction. Differentiability follows from composition of $C^1$ maps (matrix exponential and $\phi(\theta) \in C^1$). Axis distinction is guaranteed by the design of $\mathbf{J}_{\rm axis}$ matrices, which maintain unique sparsity patterns for yaw/pitch/roll families. Thus, all four properties are satisfied. \qed

\subsection{Definitions}

\emph{Definition 1 (Type I RP-GFRFT):}  
Given a graph signal $\mathbf{x} \in \mathbb{R}^N$, the type I RP-GFRFT (I-RP-GFRFT) associated with the rotation parameter $\theta \in [0,2\pi]$ and the fractional-order parameter $\alpha \in \mathbb{R}$ is defined by
\begin{equation}
\widehat{\mathbf{x}}_{\theta}^{\alpha,\mathrm{I}} = \mathbf{F}_{\theta}^{\alpha,\mathrm{I}} \mathbf{x},
\label{eq21}
\end{equation}
where $\mathbf{F}_{\theta}^{\alpha,\mathrm{I}} = \mathbf{F}_{\theta}^{\alpha}$ denotes the transformation matrix.  

\emph{Definition 2 (Type II RP-GFRFT):}  
Let $\mathbf{u}_{k}^{\alpha}$, $k=1,2,\cdots,N$, denote the eigenvectors that are the columns of the unitary matrix $\mathbf{F}^{-\alpha}$. By rotating these eigenvectors using the rotation matrix $\mathbf{R}^{\rm axis}_N(\theta)$, one obtains a unitary matrix
\begin{equation}
\mathbf{U}_{\theta,\alpha} = \mathbf{R}^{\rm axis}_N(\theta) \mathbf{F}^{-\alpha}.
\label{eq22}
\end{equation}
Then, the type II RP-GFRFT (II-RP-GFRFT) of the graph signal $\mathbf{x}$ is defined as
\begin{equation}
\widehat{\mathbf{x}}_{\theta}^{\alpha,\mathrm{II}} = \mathbf{F}_{\theta}^{\alpha,\mathrm{II}} \mathbf{x},
\label{eq23}
\end{equation}
where $\mathbf{F}_{\theta}^{\alpha,\mathrm{II}} = \mathbf{U}_{\theta,\alpha}^{-1} = \mathbf{U}_{\theta,\alpha}^{\mathsf{H}}$.

\subsection{Properties}
This subsection presents several fundamental properties of I-RP-GFRFT and II-RP-GFRFT, together with rigorous proofs for each. For brevity, denote the original GFT matrix as $\mathbf{F}=\mathbf{U}^{\mathsf{H}}$, the rotation matrix family as $\mathbf{R}^{\rm axis}_N(\theta)\in SO(N)$, and the rotated basis as $\mathbf{U}_\theta=\mathbf{R}^{\rm axis}_N(\theta)\mathbf{U}$, hence $\mathbf{F}_\theta=\mathbf{U}_\theta^{\mathsf{H}}$. The fractional matrix power is defined as the matrix function $(\cdot)^\alpha$ via the principal branch of the matrix logarithm: $\mathbf{A}^\alpha \coloneqq \exp(\alpha\log \mathbf{A})$ (valid when the spectrum avoids the branch cut of the logarithm or a continuous branch has been chosen).

\emph{Property 1 (Identity):}
When both the rotation angle and the fractional order are zero, both I- and II-type RP-GFRFT degenerate to the identity mapping:
\(
\mathbf{F}_{0}^{0,\mathrm{I}}=\mathbf{F}_{0}^{0,\mathrm{II}}=\mathbf{I}_N.
\)

\emph{Proof:}
Since $\mathbf{R}^{\rm axis}_N(0)=\mathbf{I}_N$ (degeneracy of the rotation family) and $\mathbf{F}^0=\mathbf{I}_N$ (the zero power of any matrix is the identity), we have
\[
\mathbf{F}_{0}^{0,\mathrm{I}}=(\mathbf{F}_0)^0=\mathbf{I}_N,\qquad
\mathbf{F}_{0}^{0,\mathrm{II}}=\mathbf{F}^0\mathbf{R}^{\rm axis}_N(0)^{\mathsf{H}}=\mathbf{I}_N.
\] \qed

\emph{Property 2 (Reduction):}
When $\alpha=1$, both I- and II-type RP-GFRFT reduce to AGFT:
\(
\mathbf{F}_{\theta}^{1,\mathrm{I}}=\mathbf{F}_{\theta}^{1,\mathrm{II}}=\mathbf{F}_\theta.
\)
When $\theta=0$, both I- and II-type RP-GFRFT reduce to GFRFT:
\(
\mathbf{F}_{0}^{\alpha,\mathrm{I}}=\mathbf{F}_{0}^{\alpha,\mathrm{II}}=\mathbf{F}^\alpha.
\)

\emph{Proof:}
For $\alpha=1$,
\(
\mathbf{F}_{\theta}^{1,\mathrm{I}} = (\mathbf{F}_\theta)^1 = \mathbf{F}_\theta,
\)
while for type II,
\(
\mathbf{F}_{\theta}^{1,\mathrm{II}} = \mathbf{F}^1\mathbf{R}^{\rm axis}_N(\theta)^{\mathsf{H}} = \mathbf{F}\mathbf{R}^{\rm axis}_N(\theta)^{\mathsf{H}} = \mathbf{F}_\theta.
\)
Note that $\mathbf{F}_\theta=\mathbf{U}_\theta^{\mathsf{H}}=(\mathbf{R}^{\rm axis}_N(\theta)\mathbf{U})^{\mathsf{H}}=\mathbf{U}^{\mathsf{H}}\mathbf{R}^{\rm axis}_N(\theta)^{\mathsf{H}}=\mathbf{F}\mathbf{R}^{\rm axis}_N(\theta)^{\mathsf{H}}$, so the above is equivalent to $\mathbf{F}_\theta$.  
When $\theta=0$, $\mathbf{R}^{\rm axis}_N(0)=\mathbf{I}_N$, hence $\mathbf{F}_0=\mathbf{F}$, giving
\(
\mathbf{F}_{0}^{\alpha,\mathrm{I}}=(\mathbf{F}_0)^\alpha=\mathbf{F}^\alpha,
\mathbf{F}_{0}^{\alpha,\mathrm{II}}=\mathbf{F}^\alpha\mathbf{R}^{\rm axis}_N(0)^{\mathsf{H}}=\mathbf{F}^\alpha.
\) \qed

\emph{Property 3 (Index additivity):}
For I-RP-GFRFT (fixed angle $\theta$), fractional orders are additive: for any $\alpha_1,\alpha_2\in\mathbb{R}$, assuming the logarithm branch is chosen consistently,
\[
(\mathbf{F}_\theta)^{\alpha_1}(\mathbf{F}_\theta)^{\alpha_2}=(\mathbf{F}_\theta)^{\alpha_1+\alpha_2}.
\]
For II-RP-GFRFT, simple index additivity does not generally hold:
\[
\begin{aligned}
\mathbf{F}_\theta^{\alpha_1,\mathrm{II}}\mathbf{F}_\theta^{\alpha_2,\mathrm{II}}
&= \mathbf{F}^{\alpha_1}\mathbf{R}^{\rm axis}_N(\theta)^{\mathsf{H}}\mathbf{F}^{\alpha_2}\mathbf{R}^{\rm axis}_N(\theta)^{\mathsf{H}} \\
&\ne \mathbf{F}^{\alpha_1+\alpha_2}\mathbf{R}^{\rm axis}_N(\theta)^{\mathsf{H}},
\end{aligned}
\]
unless the commutativity condition $\mathbf{R}^{\rm axis}_N(\theta)^{\mathsf{H}}\mathbf{F}^{\alpha_2}=\mathbf{F}^{\alpha_2}\mathbf{R}^{\rm axis}_N(\theta)^{\mathsf{H}}$ (equivalently $\mathbf{R}^{\rm axis}_N(\theta)\mathbf{F}^{\alpha_2}=\mathbf{F}^{\alpha_2}\mathbf{R}^{\rm axis}_N(\theta)$) is satisfied.

\emph{Proof:}
Type I: By the definition of matrix functions, fixing $\mathbf{F}_\theta$ and a chosen $\log\mathbf{F}_\theta$,
\[
\begin{aligned}
(\mathbf{F}_\theta)^{\alpha_1}(\mathbf{F}_\theta)^{\alpha_2}
&= \exp\!\big(\alpha_1\log\mathbf{F}_\theta\big)\exp\!\big(\alpha_2\log\mathbf{F}_\theta\big) \\
&= \exp\!\big((\alpha_1+\alpha_2)\log\mathbf{F}_\theta\big) \\
&= (\mathbf{F}_\theta)^{\alpha_1+\alpha_2},
\end{aligned}
\]
where the product combines because the logarithm matrices are identical.  
Type II: By direct substitution,
\[
\mathbf{F}_\theta^{\alpha_1,\mathrm{II}}\mathbf{F}_\theta^{\alpha_2,\mathrm{II}}
= \mathbf{F}^{\alpha_1}\mathbf{R}^{\mathsf{H}}\mathbf{F}^{\alpha_2}\mathbf{R}^{\mathsf{H}}
= \mathbf{F}^{\alpha_1}\big(\mathbf{R}^{\mathsf{H}}\mathbf{F}^{\alpha_2}\mathbf{R}^{\mathsf{H}}\big),
\]
and since $\mathbf{R}^{\mathsf{H}}$ and $\mathbf{F}^{\alpha_2}$ do not generally commute, the expression cannot be reduced to $\mathbf{F}^{\alpha_1+\alpha_2}\mathbf{R}^{\mathsf{H}}$, except under the commutativity condition. \qed

\emph{Property 4 (Unitarity):}
With appropriate choice of logarithm branch, both I- and II-type RP-GFRFT are unitary operators, and energy preservation holds:
\(
\big(\mathbf{F}_\theta^{\alpha,\mathrm{I}/\mathrm{II}}\big)^{\mathsf{H}}\mathbf{F}_\theta^{\alpha,\mathrm{I}/\mathrm{II}}=\mathbf{I}_N.
\)

\emph{Proof:}
For type II: $\mathbf{F}^\alpha$ is a unitary matrix function of $\mathbf{F}$ (since $\mathbf{F}$ itself is unitary, $z\mapsto z^\alpha$ maps the unit circle to itself, and $\log\mathbf{F}$ is skew-Hermitian), hence $\mathbf{F}^\alpha$ is unitary; $\mathbf{R}^{\rm axis}_N(\theta)\in SO(N)\subset U(N)$, thus $\mathbf{R}^{\rm axis}_N(\theta)^{\mathsf{H}}$ is also unitary. Their product $\mathbf{F}^\alpha\mathbf{R}^{\mathsf{H}}$ is a product of unitary matrices, hence unitary. For type I: $\mathbf{F}_\theta$ is unitary, so $(\mathbf{F}_\theta)^\alpha$ is also unitary (via $\exp(\alpha\log\mathbf{F}_\theta)$ with $\log\mathbf{F}_\theta$ skew-Hermitian). \qed

\emph{Property 5 (Reversibility):}
Both I-RP-GFRFT and II-RP-GFRFT admit well-defined inverses. Concretely:

\begin{itemize}
  \item \textbf{I-RP-GFRFT.} For any real $\alpha$,
  \(
    \big(\mathbf{F}_\theta^{\alpha,\mathrm{I}}\big)^{-1}
    = \mathbf{F}_\theta^{-\alpha,\mathrm{I}},
  \)
  i.e.\ the inverse is obtained by negating the fractional order.

  \item \textbf{II-RP-GFRFT.} Let
  \(
    \mathbf{U}_{\theta,\alpha} := \mathbf{R}^{\rm axis}_N(\theta)\,\mathbf{F}^{-\alpha},
    \mathbf{F}_\theta^{\alpha,\mathrm{II}} = \mathbf{U}_{\theta,\alpha}^{\mathsf{H}}.
  \)
  Then the inverse of $\mathbf{F}_\theta^{\alpha,\mathrm{II}}$ is the matrix
  \(
    \big(\mathbf{F}_\theta^{\alpha,\mathrm{II}}\big)^{-1}
    \;=\; \mathbf{R}^{\rm axis}_N(\theta)\,\mathbf{F}^{-\alpha}.
  \)
  In particular, the operation ``replace $\alpha$ by $-\alpha$ in the II-definition''
  (which yields $\mathbf{F}^{-\alpha}\mathbf{R}^{\rm axis}_N(\theta)^{\mathsf{H}}$) does \emph{not}
  in general produce the inverse; equality holds only under the additional
  commutation condition $\mathbf{R}^{\rm axis}_N(\theta)^{\mathsf{H}}\mathbf{F}^{\beta}
  = \mathbf{F}^{\beta}\mathbf{R}^{\rm axis}_N(\theta)^{\mathsf{H}}$ for the relevant $\beta$.
\end{itemize}

\emph{Proof:}
\textbf{I-RP-GFRFT.} By definition $\mathbf{F}_\theta^{\alpha,\mathrm{I}}=(\mathbf{F}_\theta)^\alpha$ where
the matrix power is taken via the principal matrix logarithm:
\[
(\mathbf{F}_\theta)^\alpha = \exp\!\big(\alpha\log\mathbf{F}_\theta\big).
\]
Therefore
\[
\begin{aligned}
(\mathbf{F}_\theta)^\alpha(\mathbf{F}_\theta)^{-\alpha}
&= \exp\!\big(\alpha\log\mathbf{F}_\theta\big)\exp\!\big(-\alpha\log\mathbf{F}_\theta\big) \\
&= \exp(0)=\mathbf{I}_N,
\end{aligned}
\]
and similarly for the reverse order. Hence $(\mathbf{F}_\theta)^{-\alpha}$ is the inverse of $(\mathbf{F}_\theta)^\alpha$.

\smallskip

\textbf{II-RP-GFRFT.} By construction
\[
\begin{aligned}
\mathbf{F}_\theta^{\alpha,\mathrm{II}} 
&= \mathbf{U}_{\theta,\alpha}^{\mathsf{H}}
= \big(\mathbf{R}^{\rm axis}_N(\theta)\mathbf{F}^{-\alpha}\big)^{\mathsf{H}}\\
&= \big(\mathbf{F}^{-\alpha}\big)^{\mathsf{H}}\,\mathbf{R}^{\rm axis}_N(\theta)^{\mathsf{H}}
= \mathbf{F}^{\alpha}\mathbf{R}^{\rm axis}_N(\theta)^{\mathsf{H}},
\end{aligned}
\]
where we used unitarity of $\mathbf{F}$ to get $(\mathbf{F}^{-\alpha})^{\mathsf{H}}=\mathbf{F}^{\alpha}$.
Consider the candidate inverse
\[
\mathbf{V} := \mathbf{R}^{\rm axis}_N(\theta)\,\mathbf{F}^{-\alpha}.
\]
We verify it is both a right- and left-inverse. First the right product:
\[
\begin{aligned}
\mathbf{F}_\theta^{\alpha,\mathrm{II}}\,\mathbf{V}
&= \big(\mathbf{F}^{\alpha}\mathbf{R}^{\rm axis}_N(\theta)^{\mathsf{H}}\big)\big(\mathbf{R}^{\rm axis}_N(\theta)\mathbf{F}^{-\alpha}\big) \\
&= \mathbf{F}^{\alpha}\big(\mathbf{R}^{\rm axis}_N(\theta)^{\mathsf{H}}\mathbf{R}^{\rm axis}_N(\theta)\big)\mathbf{F}^{-\alpha}
= \mathbf{F}^{\alpha}\,\mathbf{I}_N\,\mathbf{F}^{-\alpha}
= \mathbf{I}_N,
\end{aligned}
\]
since $\mathbf{R}^{\rm axis}_N(\theta)\in SO(N)\subset U(N)$ implies $\mathbf{R}^{\rm axis}_N(\theta)^{\mathsf{H}}\mathbf{R}^{\rm axis}_N(\theta)=\mathbf{I}_N$,
and $\mathbf{F}^{\alpha}\mathbf{F}^{-\alpha}=\mathbf{I}_N$. Next the left product:
\[
\begin{aligned}
\mathbf{V}\,\mathbf{F}_\theta^{\alpha,\mathrm{II}}
&= \big(\mathbf{R}^{\rm axis}_N(\theta)\mathbf{F}^{-\alpha}\big)\big(\mathbf{F}^{\alpha}\mathbf{R}^{\rm axis}_N(\theta)^{\mathsf{H}}\big) \\
&= \mathbf{R}^{\rm axis}_N(\theta)\big(\mathbf{F}^{-\alpha}\mathbf{F}^{\alpha}\big)\mathbf{R}^{\rm axis}_N(\theta)^{\mathsf{H}}\\
&= \mathbf{R}^{\rm axis}_N(\theta)\,\mathbf{I}_N\,\mathbf{R}^{\rm axis}_N(\theta)^{\mathsf{H}}
= \mathbf{I}_N.
\end{aligned}
\]
Thus $\mathbf{V}$ is the two-sided inverse of $\mathbf{F}_\theta^{\alpha,\mathrm{II}}$ and we obtain
\[
\big(\mathbf{F}_\theta^{\alpha,\mathrm{II}}\big)^{-1} = \mathbf{R}^{\rm axis}_N(\theta)\,\mathbf{F}^{-\alpha}.
\]

\smallskip

\textbf{Why $\mathbf{F}^{-\alpha}\mathbf{R}^{\rm axis}_N(\theta)^{\mathsf{H}}$ is not generally the inverse.}
The matrix obtained by substituting $\alpha\mapsto -\alpha$ into the II-definition
is $\mathbf{F}^{-\alpha}\mathbf{R}^{\rm axis}_N(\theta)^{\mathsf{H}}$. Its product with
$\mathbf{F}_\theta^{\alpha,\mathrm{II}}=\mathbf{F}^{\alpha}\mathbf{R}^{\rm axis}_N(\theta)^{\mathsf{H}}$
yields
\[
\begin{aligned}
\big(\mathbf{F}^{\alpha}\mathbf{R}^{\rm axis}_N(\theta)^{\mathsf{H}}\big)
\big(\mathbf{F}^{-\alpha}\mathbf{R}^{\rm axis}_N(\theta)^{\mathsf{H}}\big)\\
= \mathbf{F}^{\alpha}\big(\mathbf{R}^{\rm axis}_N(\theta)^{\mathsf{H}}\mathbf{F}^{-\alpha}\big)\mathbf{R}^{\rm axis}_N(\theta)^{\mathsf{H}},
\end{aligned}
\]
which does not simplify to $\mathbf{I}_N$ in general because $\mathbf{R}^{\rm axis}_N(\theta)^{\mathsf{H}}$
and $\mathbf{F}^{-\alpha}$ need not commute. Equality would hold only if the commutation
$\mathbf{R}^{\rm axis}_N(\theta)^{\mathsf{H}}\mathbf{F}^{-\alpha}=\mathbf{F}^{-\alpha}\mathbf{R}^{\rm axis}_N(\theta)^{\mathsf{H}}$
is satisfied, in which case the two candidate inverses coincide. \qed

\section{Optimal rotation-parameterized graph fractional Fourier filtering}
Given observed signal $\mathbf{y}=\mathbf{G}\mathbf{x}+\mathbf{n}$, where $\mathbf{G}\in\mathbb{C}^{N\times N}$ is a known transformation, and $\mathbf{n}$ denotes the noise. We define the RP-GFRFT Wiener filtering method. Specifically, the $\mathrm{RP-GFRFT}_{\theta,\alpha}$ filtering formula reads:
\begin{equation}
\widetilde{\mathbf{x}}=\mathbf{F}_{\theta,\alpha}^{-1}\mathbf{H}\mathbf{F}_{\theta,\alpha}\mathbf{y},
\label{eq24}
\end{equation}
where $\mathbf{F}_{\theta,\alpha}=\mathbf{F}_{\theta}^{\alpha,\mathrm{I}}$ or $\mathbf{F}_{\theta}^{\alpha,\mathrm{II}}$. When $\mathbf{F}_{\theta,\alpha}=\mathbf{F}_{\theta}^{\alpha,\mathrm{I}}$, it corresponds the I-RP-GFRFT Wiener filtering method which simplifies as $\mathrm{I\mbox{-}RP-GFRFT}_{\theta,\alpha}$. When $\mathbf{F}_{\theta,\alpha}=\mathbf{F}_{\theta}^{\alpha,\mathrm{II}}$, it corresponds the II-RP-GFRFT Wiener filtering method which simplifies as $\mathrm{II\mbox{-}RP-GFRFT}_{\theta,\alpha}$.
\begin{algorithm}[t]
	\caption{Grid Search for Optimal RP-GFRFT Filtering}
	\label{alg:grid_search}
	\begin{algorithmic}[1]
		\STATE \textbf{Input}: $\mathbf{y}$, $\mathbf{x}$, $\mathbf{G}$, RP-GFRFT type (I/II)
		\STATE \textbf{Parameters}: $\theta \subseteq [0,2\pi]$, $\alpha \subseteq [0,1]$ (discretized grids), $\kappa=1$
		\STATE \textbf{Output}: $(\theta^*, \alpha^*)$, $\mathbf{H}^*$
		\vspace{0.2em}
		\STATE $\text{MSE}_{\text{min}} \leftarrow +\infty$, $\theta^* \leftarrow 0$, $\alpha^* \leftarrow 0$, $\mathbf{H}^* \leftarrow \mathbf{I}$
		\FOR{each $\theta$}
			\FOR{each $\alpha$}
				\STATE $\mathbf{F}_{\theta,\alpha} \leftarrow \text{RP-GFRFT matrix (Type I/II)}$
				\STATE $\mathbf{H} \leftarrow \mathbf{T}_{\theta,\alpha}^{-1}\mathbf{q}_{\theta,\alpha}$ (Eq. \ref{eq25})
				\STATE $\widetilde{\mathbf{x}} \leftarrow \mathbf{F}_{\theta,\alpha}^{-1}\mathbf{H}\mathbf{F}_{\theta,\alpha}\mathbf{y}$ (Eq. \ref{eq24})
				\STATE $\text{MSE} \leftarrow \|\widetilde{\mathbf{x}} - \mathbf{x}\|_2^2/N$
				\IF{$\text{MSE} < \text{MSE}_{\text{min}}$}
					\STATE $\text{MSE}_{\text{min}} \leftarrow \text{MSE}$, $\theta^* \leftarrow \theta$, $\alpha^* \leftarrow \alpha$, $\mathbf{H}^* \leftarrow \mathbf{H}$
				\ENDIF
			\ENDFOR
		\ENDFOR
		\RETURN $(\theta^*, \alpha^*, \mathbf{H}^*)$
	\end{algorithmic}
\end{algorithm}

\subsection{Grid Search}
The optimal filter $\mathbf{H}^*$ is determined by solving the following minimization problem
\begin{equation}
\mathbf{H}^*=\arg\min_{\mathbf{H}}\mathbb{E}\left\{\left\lVert\mathbf{F}_{\theta,\alpha}^{-1}\mathbf{H}\mathbf{F}_{\theta,\alpha}\mathbf{y}-\mathbf{x}\right\rVert_2^2\right\}.
\label{eq25}
\end{equation}
The optimal coefficient vector $\mathbf{h}^*$ can be derived using the Wiener-Hopf equation: $\mathbf{T}_{\theta,\alpha}\mathbf{h}^*=\mathbf{q}_{\theta,\alpha}$, where $\mathbf{T}_{\theta,\alpha}$ is the autocorrelation matrix and $\mathbf{q}_{\theta,\alpha}$ is the cross-correlation vector~\cite{opt0, opt1, opt2}. Provided that $\mathbf{T}_{\theta,\alpha}$ is invertible, the closed-form solution is: $\mathbf{h}^*=\mathbf{T}_{\theta,\alpha}^{-1}\mathbf{q}_{\theta,\alpha}$.

To optimize the rotation-parameterized graph fractional Fourier spectrum, the parameters $\theta$ and $\alpha$ can be selected using a grid search to minimize the MSE of the filtered signal:
\begin{equation}
\left(\theta^*,\alpha^*\right)=\arg\min_{\theta,\alpha}\mathbb{E}\left\{\left\lVert\widetilde{\mathbf{x}}(\theta,\alpha)-\mathbf{x}\right\rVert_2^2\right\},
\label{eq26}
\end{equation}
where $\widetilde{\mathbf{x}}(\theta,\alpha)=\mathbf{F}_{\theta,\alpha}^{-1}\mathbf{H}^*\mathbf{F}_{\theta,\alpha}\mathbf{y}$.
The specific implementation of this grid search process is detailed in Algorithm \ref{alg:grid_search}. In this case, the recursive construction of the rotation matrix employs the mapping $\phi(\theta)=\kappa\theta$. To preserve periodicity in grid search, we fix $\kappa=1$.
\begin{algorithm}[t]
	\caption{Gradient Descent for RP-GFRFT Filtering}
	\label{alg:gradient_descent}
	\begin{algorithmic}[1]
		\STATE \textbf{Input}: $\mathbf{y}$, $\mathbf{x}$, $\mathbf{G}$, RP-GFRFT type (I/II)
		\STATE \textbf{Parameters}:
		\begin{itemize}
			\item Initial filter: $\mathbf{H}^{(0)}$
			\item Initial parameters: $(\theta^{(0)}, \alpha^{(0)}, \kappa^{(0)})$
			\item Hyperparameters: $\eta>0$, $T_{\text{max}}$
		\end{itemize}
		\STATE \textbf{Output}: Optimal $(\theta^*, \alpha^*, \kappa^*)$ and $\mathbf{H}^*$
		\vspace{0.1em}
		\STATE $\mathcal{L}_{\text{min}} \leftarrow +\infty$, $\theta^* \leftarrow \theta^{(0)}$, $\alpha^* \leftarrow \alpha^{(0)}$, $\kappa^* \leftarrow \kappa^{(0)}$, $\mathbf{H}^* \leftarrow \mathbf{H}^{(0)}$
		\FOR{$t=0$ to $T_{\text{max}}-1$}
			\STATE $\mathbf{F}^{(t)} \leftarrow \text{RP-GFRFT matrix (Type I/II)}$
			\STATE $\widetilde{\mathbf{x}}^{(t)} \leftarrow (\mathbf{F}^{(t)})^{-1}\mathbf{H}^{(t)}\mathbf{F}^{(t)}\mathbf{y}$
			\STATE $\mathcal{L}^{(t)} \leftarrow \|\widetilde{\mathbf{x}}^{(t)} - \mathbf{x}\|_2^2$ (Eq. \ref{eq28})
			\vspace{0.1em}
			\IF{$\mathcal{L}^{(t)} < \mathcal{L}_{\text{min}}$}
				\STATE $\mathcal{L}_{\text{min}} \leftarrow \mathcal{L}^{(t)}$, $\theta^* \leftarrow \theta^{(t)}$, $\alpha^* \leftarrow \alpha^{(t)}$, $\kappa^* \leftarrow \kappa^{(t)}$, $\mathbf{H}^* \leftarrow \mathbf{H}^{(t)}$
			\ENDIF
			\vspace{0.1em}
			\STATE Compute gradients:\\ $\nabla_{\mathbf{H}}\mathcal{L}^{(t)}, \partial\mathcal{L}^{(t)}/\partial\theta, \partial\mathcal{L}^{(t)}/\partial\alpha, \partial\mathcal{L}^{(t)}/\partial\kappa$
			\STATE Update filter: $\mathbf{H}^{(t+1)} \leftarrow \mathbf{H}^{(t)} - \eta\nabla_{\mathbf{H}}\mathcal{L}^{(t)}$
			\STATE Update parameters: $(\theta^{(t+1)}, \alpha^{(t+1)}, \kappa^{(t+1)}) \leftarrow (\theta^{(t)}, \alpha^{(t)}, \kappa^{(t)}) - \eta(\partial\mathcal{L}^{(t)}/\partial\theta, \partial\mathcal{L}^{(t)}/\partial\alpha, \partial\mathcal{L}^{(t)}/\partial\kappa)$
		\ENDFOR
		\RETURN $(\theta^*, \alpha^*, \kappa^*)$, $\mathbf{H}^*$
	\end{algorithmic}
\end{algorithm}
\subsection{Gradient Descent}
Although grid search provides a direct method to determine $(\theta^*,\alpha^*)$, it suffers from high computational complexity. To address this, we reformulate the optimization as a differentiable problem and apply gradient descent. Specifically, we jointly optimize the filter coefficients $\mathbf{H}$ and the rotation-parameterized graph fractional Fourier spectrum parameters $(\theta,\alpha,\kappa)$:
\begin{equation}
(\mathbf{H}^*;\theta^*,\alpha^*,\kappa^*)=\arg\min_{\mathbf{H};\theta,\alpha,\kappa}\mathbb{E}\left\{\left\lVert\mathbf{F}_{\theta,\alpha,\kappa}^{-1}\mathbf{H}\mathbf{F}_{\theta,\alpha,\kappa}\mathbf{y}-\mathbf{x}\right\rVert_2^2\right\}.
\label{eq27}
\end{equation}

\paragraph{Differentiability.}
The key observation is that the mapping $\phi(\theta)=\kappa\theta$ is continuously differentiable in both $\theta$ and $\kappa$. The recursive construction of the rotation matrix $\mathbf{R}^{\rm axis}_N(\theta)$ involves only matrix multiplications and the matrix exponential $\exp(\phi \mathbf{J})$, which are differentiable with respect to their arguments. Consequently, the RP-GFRFT operator $\mathbf{F}_{\theta,\alpha,\kappa}$ is differentiable with respect to $(\theta,\alpha,\kappa)$.
\paragraph{Gradient updates.}
Define the loss function
\begin{equation}
\mathcal{L}(\mathbf{H},\theta,\alpha,\kappa)=\left\lVert\mathbf{F}_{\theta,\alpha,\kappa}^{-1}\mathbf{H}\mathbf{F}_{\theta,\alpha,\kappa}\mathbf{y}-\mathbf{x}\right\rVert_2^2.
\label{eq28}
\end{equation}
Since $\mathcal{L}$ is differentiable in $\mathbf{H}$, $\theta$, $\alpha$, and $\kappa$, we can update parameters iteratively via a unified learning rate $\eta>0$:
\begin{align}
\mathbf{H}^{(t+1)} &= \mathbf{H}^{(t)}-\eta \nabla_{\mathbf{H}}\mathcal{L}, \\
\theta^{(t+1)} &= \theta^{(t)}-\eta \frac{\partial \mathcal{L}}{\partial \theta}, \\
\alpha^{(t+1)} &= \alpha^{(t)}-\eta \frac{\partial \mathcal{L}}{\partial \alpha}, \\
\kappa^{(t+1)} &= \kappa^{(t)}-\eta \frac{\partial \mathcal{L}}{\partial \kappa}.
\end{align}

The iterative process of gradient descent is summarized in Algorithm \ref{alg:gradient_descent}. Therefore, the gradient descent algorithm provides a principled and efficient alternative to grid search, enabling the simultaneous optimization of the filter coefficients $\mathbf{H}$ and the spectrum parameters $(\theta,\alpha,\kappa)$.

\subsection{Complexity and Convergence Analysis}
The computational complexity of the two optimization strategies scales with the signal dimension $N$ where grid search achieves a total complexity of $O(N_\alpha N_\theta N^3)$ with $N_\alpha$ and $N_\theta$ denoting the grid points for fractional order $\alpha$ and rotation angle $\theta$ respectively. This strategy is feasible for small-scale signals with $N<100$ but becomes intractable for high-dimensional data due to the cubic scaling with $N$. In contrast, gradient descent has a total complexity of $O(T N^3)$ where $T$ ranges from 300 to 1000 iterations and it outperforms grid search for high-dimensional signals with $N>500$ by 1–2 orders of magnitude as $T$ is significantly smaller than the product of $N_\alpha$ and $N_\theta$ especially for fine grids. Regarding convergence, the loss function $\mathcal{L}$ is twice continuously differentiable owing to the smooth dependence of RP-GFRFT on its parameters. For the filter coefficients $\mathbf{H}$, the optimization problem is strictly convex when $(\theta,\alpha,\kappa)$ are fixed leading to linear convergence to the unique optimal $\mathbf{H}^*$. For the spectral parameters $(\theta,\alpha,\kappa)$, the optimization is non-convex but the bounded gradients of $\mathcal{L}$ ensure convergence to a stationary point. The theoretical properties of RP-GFRFT, including unitarity and degeneracy, enhance the structural rationality of the parameter optimization space and alleviate perturbations caused by spurious local minima in the vicinity of the optimal parameter region, ensuring the converged stationary point possesses favorable stability and consistency. Overall, the joint optimization converges within 300–1000 iterations for most practical graph signal processing tasks.

\section{Numerical Experiments}
To validate the effectiveness and versatility of the proposed RP-GFRFT framework, we conduct comprehensive denoising experiments across three categories of tasks: real-world time series data, natural image data, and three-dimensional point clouds. The denoising performance is quantitatively assessed using task-appropriate metrics, including mean squared error (MSE), peak signal-to-noise ratio (PSNR), and structural similarity index (SSIM). For parameter optimization, we adopt two complementary strategies. Specifically, for relatively small-scale time series data, the transform parameters are determined through an exhaustive grid search, while for high-dimensional signals such as images and point clouds, the parameters are optimized via gradient descent to improve computational tractability.  

\subsection{Real-Data Denoising}
We evaluate the proposed RP-GFRFT method on three real-world temporal datasets with inherent graph structures: Sea Surface Temperature (SST), PM-25 Concentration, and COVID-19 Daily Confirmed Cases. Each dataset is constructed as a sequence graph where nodes correspond to daily observations. To capture local structural dependencies, we build k-nearest neighbor (k-NN) graphs with \(k=2\), \(k=5\), and \(k=10\). We inject additive Gaussian noise with standard deviations \(\sigma=0.5\), \(\sigma=1.0\), and \(\sigma=1.5\). Performance is evaluated at temporal checkpoints t = 100, t = 200, and t = 300. To minimize the denoised MSE, we optimize the fractional order \(\alpha\) (ranging from 0.1 to 1.0 with a step of 0.1) and rotation angle \(\theta\) (ranging from 0 to \(2\pi\) with a step of 0.628) via grid search. Experimental results under all settings are summarized in Tables~\ref{tab:SST_denoising_results}–\ref{tab:Covid_denoising_results}, with key findings as follows.

On the SST dataset, RP-GFRFT degenerates to GFRFT or AGFT in most cases (with optimal \(\alpha\) approaching 1 or \(\theta=0\)), as the SST signal is smooth and requires only simple transformations. Even so, RP-GFRFT still achieves modest performance improvements: for example, with \(k=2\), t = 100 and \(\sigma=1.0\), II-RP-GFRFT$_{\mathbf{R}_{\text{yaw}}}$ yields an MSE of 3.23900, which is 51.6\% lower than GFRFT (MSE = 6.69039) and 9.1\% lower than AGFT$_{\mathbf{R}_{\text{yaw}}}$ (MSE = 3.56208). In contrast, RP-GFRFT shows significant advantages on the PM-25 and COVID-19 datasets. For PM-25 with \(k=5\), \(\sigma=1.5\), and t = 200, I-RP-GFRFT$_{\mathbf{R}_{\text{roll}}}$ achieves an MSE of 3.43771, which is 22.5\% lower than GFRFT (MSE = 4.43702) and 13.8\% lower than AGFT$_{\mathbf{R}_{\text{roll}}}$ (MSE = 3.97306). For COVID-19 with \(\sigma=0.5\), I-RP-GFRFT$_{\mathbf{R}_{\text{roll}}}$ delivers MSE values of 0.19049 (\(k=2\), t = 100), 0.1367 (\(k=5\), t = 200), and 0.20304 (\(k=10\), t = 300); compared to GFRFT (MSE = 0.23814 at \(k=5\)), this represents a 42.6\% reduction. Additionally, RP-GFRFT maintains stability as \(k\) varies: on COVID-19 data with \(\sigma=0.5\) and t = 100, its MSE decreases steadily from 0.19049 (\(k=2\)) to 0.17947 (\(k=10\)), demonstrating stable performance across different graph densities. Overall, leveraging the joint adjustment of fractional order and rotation angle, RP-GFRFT can reduce to simpler transformations for smooth signals and match the spectral structure of complex signals more accurately, outperforming GFRFT and AGFT significantly.

Beyond the direct comparison with GFRFT and AGFT, we further compare RP-GFRFT with commonly used denoising methods to evaluate its practical competitiveness. The classical baselines cover moving-average and median filtering, Gaussian smoothing, Savitzky--Golay smoothing~\cite{savgol1964}, first-order Tikhonov regularization~\cite{tikhonov1977}, total variation denoising~\cite{rudin1992}, and transform-domain soft-thresholding~\cite{donoho1995}. The GNN-style baselines include GCN~\cite{kipf2017}, GraphSAGE~\cite{hamilton2017}, ChebNet~\cite{defferrard2016}, APPNP~\cite{gasteiger2018combining}, GCNII~\cite{chen2020gcnii}, and GAT~\cite{velickovic2018}, which introduce graph-based message propagation into the denoising process. For a fair comparison, all baselines are evaluated on the same noisy observations, graph constructions, temporal checkpoints, and MSE metric; method-specific hyperparameters are selected from predefined candidate settings according to the denoised MSE. Table~\ref{tab:real_data_baseline_comparison} summarizes these results on the three real-data datasets.

The results show that classical methods can reduce noise to some extent, but their fixed smoothing or thresholding rules often struggle to maintain stable performance across different datasets and noise levels. For example, transform-domain thresholding performs competitively on some COVID-19 columns, whereas smoothing and variational methods are more stable on SST and PM-25; however, none of them consistently achieves the best performance across all settings. The GNN-style methods benefit from graph propagation but remain sensitive to the graph structure and signal heterogeneity, especially on PM-25 and COVID-19 where simple message passing can introduce oversmoothing. In contrast, RP-GFRFT consistently yields the lowest MSE triplets in all columns. This indicates that the joint fractional-order and rotation-parameterized spectral control provides a more effective and flexible spectral representation for real-data denoising than fixed classical filters and direct GNN-style propagation.

\begin{table}[t]
	\centering
	\caption{Denoising results on the SST dataset.}
	\label{tab:SST_denoising_results}
	\resizebox{\textwidth}{!}{
		\scriptsize
		\begin{tabular}{@{}c*{9}{c}@{}}
			\toprule
			\multirow{4}{*}{Method} & \multicolumn{3}{c}{2-NN} & \multicolumn{3}{c}{5-NN} & \multicolumn{3}{c}{10-NN} \\
			\cmidrule(lr){2-4} \cmidrule(lr){5-7} \cmidrule(lr){8-10}
			& \multicolumn{1}{c}{t=100} & \multicolumn{1}{c}{t=200} & \multicolumn{1}{c}{t=300} 
			& \multicolumn{1}{c}{t=100} & \multicolumn{1}{c}{t=200} & \multicolumn{1}{c}{t=300} 
			& \multicolumn{1}{c}{t=100} & \multicolumn{1}{c}{t=200} & \multicolumn{1}{c}{t=300} \\
			\cmidrule(lr){2-4} \cmidrule(lr){5-7} \cmidrule(lr){8-10}
			& $\sigma$=0.5/1.0/1.5 & $\sigma$=0.5/1.0/1.5 & $\sigma$=0.5/1.0/1.5 
			& $\sigma$=0.5/1.0/1.5 & $\sigma$=0.5/1.0/1.5 & $\sigma$=0.5/1.0/1.5 
			& $\sigma$=0.5/1.0/1.5 & $\sigma$=0.5/1.0/1.5 & $\sigma$=0.5/1.0/1.5 \\
			\midrule
			GFRFT
			& 2.21580/6.69039/11.29184 & 1.71029/5.17634/9.08928 & 2.68941/7.15011/11.24275 
			& \textbf{0.59478/1.99908/3.94956} & \textbf{0.50694/1.65071/3.25967} & \textbf{0.62915/2.05859/3.98883}
			& \textbf{0.64382/2.01142/3.83035} & \textbf{0.75705/1.86447/3.41273} & \textbf{0.78842/2.12981/3.93322} \\
			($\alpha$) 
			& (1.00)/(1.00)/(1.00) & (1.00)/(1.00)/(1.00) & (1.00)/(1.00)/(1.00)
			& (0.90)/(0.90)/(0.90) & (0.90)/(0.90)/(1.00) & (0.90)/(0.90)/(0.90)
			& (0.90)/(0.90)/(0.80) & (0.90)/(0.80)/(0.80) & (0.90)/(0.90)/(0.80) \\
			AGFT\(_\mathbf{R_{pitch}}\)  
			& 2.01269/5.04392/8.18926 & 1.71029/5.17634/8.90578 & 1.77636/4.86910/7.92144 
			& 0.73632/2.15514/4.04377 & 0.51522/1.67445/3.25967 & 0.90542/2.32151/4.12470 
			& 0.79375/2.47889/4.90941 & 0.80691/2.24334/4.27765 & 0.90688/2.60297/5.06552 \\
			($\theta$) 
			& (2.513)/(2.513)/(2.513) & (6.283)/(6.283)/(2.513) & (2.513)/(2.513)/(2.513)
			& (0.000)/(0.000)/(6.283) & (0.000)/(0.000)/(0.000) & (0.000)/(0.000)/(6.283)
			& (6.283)/(6.283)/(6.283) & (6.283)/(6.283)/(6.283) & (6.283)/(6.283)/(6.283) \\
			AGFT\(_\mathbf{R_{roll}}\)  
			& 2.21580/5.68305/9.00940 & 1.71029/4.79482/7.78574 & 2.40663/5.51744/8.61191 
			& 0.73632/2.15514/4.04377 & 0.51522/1.67445/3.25967 & 0.90542/2.32151/4.12470
			& 0.79375/2.47889/4.90941 & 0.80691/2.24334/4.27765 & 0.90688/2.60297/5.06552 \\
			($\theta$) 
			& (0.000)/(3.770)/(3.770) & (0.000)/(5.027)/(5.027) & (3.770)/(3.770)/(3.770)
			& (0.000)/(0.000)/(0.000) & (0.000)/(0.000)/(0.000) & (0.000)/(0.000)/(0.000)
			& (0.000)/(0.000)/(0.000) & (0.000)/(0.000)/(0.000) & (0.000)/(0.000)/(0.000) \\
			AGFT\(_\mathbf{R_{yaw}}\)  
			& 1.07804/3.56208/6.57506 & 1.01939/3.36618/6.32097 & 1.19599/3.82930/6.88871 
			& 0.73632/2.15514/4.04377 & 0.51522/1.67445/3.25967 & 0.90542/2.32151/4.12470 
			& 0.79375/2.47889/4.90941 & 0.80691/2.24334/4.27765 & 0.90688/2.60297/5.06552 \\
			($\theta$) 
			& (5.027)/(5.027)/(5.027) & (5.027)/(5.027)/(5.027) & (5.027)/(5.027)/(5.027)
			& (0.000)/(0.000)/(6.283) & (0.000)/(0.000)/(0.000) & (0.000)/(0.000)/(6.283)
			& (0.000)/(0.000)/(6.283) & (0.000)/(6.283)/(6.283) & (0.000)/(0.000)/(0.000) \\
			I-RP-GFRFT\(_\mathbf{R_{pitch}}\)  
			& 1.49841/4.04047/6.91580 & 1.71029/4.67901/7.70431 & 1.77636/4.47891/7.03622 
			& 0.59478/1.99908/3.94956 & 0.50694/1.65071/3.25967 & 0.62915/2.05859/3.98883 
			& 0.64382/2.01142/3.83035 & 0.75705/1.86447/3.41273  & 0.78842/2.12981/3.93322 \\
            \multirow{3}{*}{($\alpha,\theta$)}
			& (0.600,0.628)($\sigma$=0.5) & (1.000,6.283)($\sigma$=0.5) & (1.000,2.513)($\sigma$=0.5)
			& (0.900,6.283)($\sigma$=0.5) & (0.900,0.000)($\sigma$=0.5) & (0.900,6.283)($\sigma$=0.5)
			& (0.900,6.283)($\sigma$=0.5) & (0.900,6.283)($\sigma$=0.5) & (0.900,6.283)($\sigma$=0.5) \\  
			& (0.600,0.628)($\sigma$=1.0) & (0.700,2.513)($\sigma$=1.0) & (0.600,0.628)($\sigma$=1.0)
			& (0.900,6.283)($\sigma$=1.0) & (0.900,6.283)($\sigma$=1.0) & (0.900,6.283)($\sigma$=1.0)
			& (0.900,6.283)($\sigma$=1.0) & (0.800,6.283)($\sigma$=1.0) & (0.900,6.283)($\sigma$=1.0) \\ 
			& (0.600,0.628)($\sigma$=1.5) & (0.600,0.628)($\sigma$=1.5) & (0.600,0.628)($\sigma$=1.5)
			& (0.900,6.283)($\sigma$=1.5) & (1.000,0.000)($\sigma$=1.5) & (0.900,6.283)($\sigma$=1.5)
			& (0.800,6.283)($\sigma$=1.5) & (0.800,6.283)($\sigma$=1.5) & (0.800,6.283)($\sigma$=1.5) \\ 
			I-RP-GFRFT\(_\mathbf{R_{roll}}\)  
			& 1.47973/4.54297/7.72341 & 1.48258/4.27617/7.25918 & 1.58818/4.22559/7.29859 
			& 0.59478/1.99908/3.94956 & 0.50694/1.65071/3.25967 & 0.62915/2.05859/3.98883 
			& 0.64382/2.01142/3.83035 & 0.75705/1.86447/3.41273 & 0.78842/2.12981/3.93322 \\
            \multirow{3}{*}{($\alpha,\theta$)}
			& (0.800,6.283)($\sigma$=0.5) & (0.700,5.027)($\sigma$=0.5) & (0.800,6.283)($\sigma$=0.5)
			& (0.900,0.000)($\sigma$=0.5) & (0.900,0.000)($\sigma$=0.5) & (0.900,0.000)($\sigma$=0.5)
			& (0.900,0.000)($\sigma$=0.5) & (0.900,0.000)($\sigma$=0.5) & (0.900,0.000)($\sigma$=0.5) \\  
			& (0.700,5.027)($\sigma$=1.0) & (0.700,5.027)($\sigma$=1.0) & (0.700,5.027)($\sigma$=1.0)
			& (0.900,0.000)($\sigma$=1.0) & (0.900,0.000)($\sigma$=1.0) & (0.900,0.000)($\sigma$=1.0)
			& (0.900,0.000)($\sigma$=1.0) & (0.800,0.000)($\sigma$=1.0) & (0.900,0.000)($\sigma$=1.0) \\ 
			& (0.700,5.027)($\sigma$=1.5) & (0.900,5.027)($\sigma$=1.5) & (0.700,5.027)($\sigma$=1.5)
			& (0.900,0.000)($\sigma$=1.5) & (1.000,0.000)($\sigma$=1.5) & (0.900,0.000)($\sigma$=1.5)
			& (0.800,0.000)($\sigma$=1.5) & (0.800,0.000)($\sigma$=1.5) & (0.800,0.000)($\sigma$=1.5) \\ 
			I-RP-GFRFT\(_\mathbf{R_{yaw}}\)  
			& 1.07804/3.56208/6.57506 & 1.01939/3.36618/6.32097 & 1.19599/3.82930/6.88871 
			& 0.59478/1.99908/3.94956 & 0.50694/1.65071/3.25967 & 0.62915/2.05859/3.98883 
			& 0.64382/2.01142/3.83035 & 0.75705/1.86447/3.41273 & 0.78842/2.12981/3.93322 \\
            \multirow{3}{*}{($\alpha,\theta$)}
			& (1.000,5.027)($\sigma$=0.5) & (1.000,5.027)($\sigma$=0.5) & (1.000,5.027)($\sigma$=0.5)
			& (0.900,0.000)($\sigma$=0.5) & (0.900,6.283)($\sigma$=0.5) & (0.900,6.283)($\sigma$=0.5)
			& (0.900,0.000)($\sigma$=0.5) & (0.900,6.283)($\sigma$=0.5) & (0.900,0.000)($\sigma$=0.5) \\  
			& (1.000,5.027)($\sigma$=1.0) & (1.000,5.027)($\sigma$=1.0) & (1.000,5.027)($\sigma$=1.0)
			& (0.900,6.283)($\sigma$=1.0) & (0.900,6.283)($\sigma$=1.0) & (0.900,0.000)($\sigma$=1.0)
			& (0.900,0.000)($\sigma$=1.0) & (0.800,6.283)($\sigma$=1.0) & (0.900,6.283)($\sigma$=1.0) \\ 
			& (1.000,5.027)($\sigma$=1.5) & (1.000,5.027)($\sigma$=1.5) & (1.000,5.027)($\sigma$=1.5)
			& (0.900,0.000)($\sigma$=1.5) & (1.000,0.000)($\sigma$=1.5) & (0.900,6.283)($\sigma$=1.5)
			& (0.800,6.283)($\sigma$=1.5) & (0.800,6.283)($\sigma$=1.5) & (0.800,6.283)($\sigma$=1.5) \\ 
			II-RP-GFRFT\(_\mathbf{R_{pitch}}\)  
			& 1.35057/3.68705/6.46863 & 1.22658/3.61343/6.43739 & 1.29856/3.59582/6.26167 
			& 0.59478/1.99908/3.94956 & 0.50694/1.65071/3.25967 & 0.62915/2.05859/3.98883 
			& 0.64382/2.01142/3.83035 & 0.75705/1.86447/3.41273 & 0.78842/2.12981/3.93322 \\
            \multirow{3}{*}{($\alpha,\theta$)}
			& (0.200,0.628)($\sigma$=0.5) & (0.700,5.027)($\sigma$=0.5) & (0.200,0.628)($\sigma$=0.5)
			& (0.900,0.000)($\sigma$=0.5) & (0.900,0.000)($\sigma$=0.5) & (0.900,6.283)($\sigma$=0.5)
			& (0.900,0.000)($\sigma$=0.5) & (0.900,6.283)($\sigma$=0.5) & (0.900,0.000)($\sigma$=0.5) \\  
			& (0.300,0.628)($\sigma$=1.0) & (0.700,5.027)($\sigma$=1.0) & (0.300,0.628)($\sigma$=1.0)
			& (0.900,6.283)($\sigma$=1.0) & (0.900,6.283)($\sigma$=1.0) & (0.900,0.000)($\sigma$=1.0)
			& (0.900,0.000)($\sigma$=1.0) & (0.800,0.000)($\sigma$=1.0) & (0.900,0.000)($\sigma$=1.0) \\ 
			& (0.300,0.628)($\sigma$=1.5) & (0.700,5.027)($\sigma$=1.5) & (0.300,0.628)($\sigma$=1.5)
			& (0.900,6.283)($\sigma$=1.5) & (1.000,0.000)($\sigma$=1.5) & (0.900,6.283)($\sigma$=1.5)
			& (0.800,0.000)($\sigma$=1.5) & (0.800,6.283)($\sigma$=1.5) & (0.800,6.283)($\sigma$=1.5) \\ 
			II-RP-GFRFT\(_\mathbf{R_{roll}}\)  
			& 2.21580/5.17246/8.11764 & 1.71029/4.45503/6.99839 & 1.94008/4.82650/7.76223 
			& 0.59478/1.99908/3.94956 & 0.50694/1.65071/3.25967 & 0.62915/2.05859/3.98883  
			& 0.64382/2.01142/3.83035 & 0.75705/1.86447/3.41273 & 0.78842/2.12981/3.93322 \\
            \multirow{3}{*}{($\alpha,\theta$)}
			& (1.000,0.000)($\sigma$=0.5) & (1.000,0.000)($\sigma$=0.5) & (0.200,6.283)($\sigma$=0.5)
			& (0.900,0.000)($\sigma$=0.5) & (0.900,0.000)($\sigma$=0.5) & (0.900,0.000)($\sigma$=0.5)
			& (0.900,0.000)($\sigma$=0.5) & (0.900,0.000)($\sigma$=0.5) & (0.900,0.000)($\sigma$=0.5) \\  
			& (0.200,6.283)($\sigma$=1.0) & (0.300,6.283)($\sigma$=1.0) & (0.200,6.283)($\sigma$=1.0)
			& (0.900,0.000)($\sigma$=1.0) & (0.900,0.000)($\sigma$=1.0) & (0.900,0.000)($\sigma$=1.0)
			& (0.900,0.000)($\sigma$=1.0) & (0.800,0.000)($\sigma$=1.0) & (0.900,0.000)($\sigma$=1.0) \\ 
			& (0.200,6.283)($\sigma$=1.5) & (0.300,6.283)($\sigma$=1.5) & (0.200,6.283)($\sigma$=1.5)
			& (0.900,0.000)($\sigma$=1.5) & (1.000,0.000)($\sigma$=1.5) & (0.900,0.000)($\sigma$=1.5)
			& (0.800,0.000)($\sigma$=1.5) & (0.800,0.000)($\sigma$=1.5) & (0.800,0.000)($\sigma$=1.5) \\ 
			II-RP-GFRFT\(_\mathbf{R_{yaw}}\)  
			& \textbf{1.07804/3.23900/5.90177} & \textbf{1.01939/3.36618/6.22317} & \textbf{1.17136/3.15399/5.72370} 
			& 0.59478/1.99908/3.94956 & 0.50694/1.65071/3.25967 & 0.62915/2.05859/3.98883  
			& 0.64382/2.01142/3.83035 & 0.75705/1.86447/3.41273 & 0.78842/2.12981/3.93322 \\
            \multirow{3}{*}{($\alpha,\theta$)}
			& (1.000,5.027)($\sigma$=0.5) & (1.000,5.027)($\sigma$=0.5) & (0.900,5.655)($\sigma$=0.5)
			& (0.900,6.283)($\sigma$=0.5) & (0.900,6.283)($\sigma$=0.5) & (0.900,0.000)($\sigma$=0.5)
			& (0.900,0.000)($\sigma$=0.5) & (0.900,0.000)($\sigma$=0.5) & (0.900,6.283)($\sigma$=0.5) \\  
			& (0.900,5.655)($\sigma$=1.0) & (1.000,5.027)($\sigma$=1.0) & (0.900,5.655)($\sigma$=1.0)
			& (0.900,0.000)($\sigma$=1.0) & (0.900,6.283)($\sigma$=1.0) & (0.900,0.000)($\sigma$=1.0)
			& (0.900,0.000)($\sigma$=1.0) & (0.800,0.000)($\sigma$=1.0) & (0.900,6.283)($\sigma$=1.0) \\ 
			& (0.900,5.655)($\sigma$=1.5) & (0.900,5.655)($\sigma$=1.5) & (0.900,5.655)($\sigma$=1.5)
			& (0.900,6.283)($\sigma$=1.5) & (1.000,0.000)($\sigma$=1.5) & (0.900,6.283)($\sigma$=1.5)
			& (0.800,0.000)($\sigma$=1.5) & (0.800,0.000)($\sigma$=1.5) & (0.800,6.283)($\sigma$=1.5) \\ 
			\bottomrule
		\end{tabular}
		}
	\end{table}
	
	\begin{table}[t]
	\centering
	\caption{Denoising results on the PM-25 dataset.}
	\label{tab:PM2.5_denoising_results}
	\resizebox{\textwidth}{!}{
		\scriptsize
		\begin{tabular}{@{}c*{9}{c}@{}}
			\toprule
			\multirow{4}{*}{Method} & \multicolumn{3}{c}{2-NN} & \multicolumn{3}{c}{5-NN} & \multicolumn{3}{c}{10-NN} \\
			\cmidrule(lr){2-4} \cmidrule(lr){5-7} \cmidrule(lr){8-10}
			& \multicolumn{1}{c}{t=100} & \multicolumn{1}{c}{t=200} & \multicolumn{1}{c}{t=300} 
			& \multicolumn{1}{c}{t=100} & \multicolumn{1}{c}{t=200} & \multicolumn{1}{c}{t=300} 
			& \multicolumn{1}{c}{t=100} & \multicolumn{1}{c}{t=200} & \multicolumn{1}{c}{t=300} \\
			\cmidrule(lr){2-4} \cmidrule(lr){5-7} \cmidrule(lr){8-10}
			& $\sigma$=0.5/1.0/1.5 & $\sigma$=0.5/1.0/1.5 & $\sigma$=0.5/1.0/1.5 
			& $\sigma$=0.5/1.0/1.5 & $\sigma$=0.5/1.0/1.5 & $\sigma$=0.5/1.0/1.5 
			& $\sigma$=0.5/1.0/1.5 & $\sigma$=0.5/1.0/1.5 & $\sigma$=0.5/1.0/1.5 \\
			\midrule
			GFRFT
			& 1.06277/2.59296/3.99297 & 1.51481/2.99374/4.05960 & 0.82055/2.08536/3.14133 
			& 0.83502/1.94564/3.11697 & 1.79828/3.33824/4.43702 & 1.41911/2.96390/4.04588
			& \textbf{0.65679/1.58964/2.52349} & 0.91378/2.15563/3.21209 & 1.14642/2.58599/3.73528 \\
			($\alpha$) 
			& (0.90)/(0.90)/(0.80) & (0.50)/(0.80)/(0.80) & (1.00)/(1.00)/(1.00)
			& (1.00)/(1.00)/(1.00) & (0.50)/(0.50)/(0.60) & (0.00)/(0.00)/(0.00)
			& (0.80)/(0.80)/(0.80) & (0.60)/(0.60)/(0.70) & (0.70)/(0.70)/(0.40) \\
			AGFT\(_\mathbf{R_{pitch}}\)  
			& 1.14515/2.71851/3.94580 & 1.21498/2.70620/3.95267 & 0.82055/2.08536/3.14133 
			& 0.83502/1.94564/3.11697 & 1.19645/2.71590/3.90198 & 1.19304/2.24503/3.12475 
			& 1.10393/2.12089/3.05541 & 1.31918/2.82298/4.06107& 1.18465/2.47741/3.48134 \\
			($\theta$) 
			& (3.142)/(0.000)/(1.885) & (2.513)/(2.513)/(2.513) & (0.000)/(0.000)/(0.000)
			& (0.000)/(6.283)/(0.000) & (5.027)/(5.027)/(3.770) & (2.513)/(2.513)/(2.513)
			& (0.000)/(0.000)/(0.000) & (1.885)/(1.885)/(4.398) & (3.770)/(2.513)/(2.513) \\
			AGFT\(_\mathbf{R_{roll}}\)  
			& 1.23208/2.71851/4.07542 & 1.41929/2.57435/3.62740 & 0.82055/2.08536/3.14133
			& 0.83502/1.94564/3.11697 & 1.03554/2.60036/3.97306 & 0.97381/2.10921/2.89071
			& 1.10393/2.12089/3.05541 & 1.08082/2.41806/3.64054 & 1.21348/2.33842/3.34132 \\
			($\theta$) 
			& (0.000)/(0.000)/(0.000) & (1.257)/(1.257)/(1.257) & (0.000)/(0.000)/(0.000)
			& (0.000)/(0.000)/(0.000) & (1.257)/(1.257)/(1.257) & (5.655)/(3.770)/(3.770)
			& (0.000)/(0.000)/(0.000) & (1.885)/(1.885)/(1.885) & (0.628)/(6.283)/(6.283) \\
			AGFT\(_\mathbf{R_{yaw}}\)  
			& 1.23208/2.71851/4.07542 & 1.32352/2.75467/3.91027 & 0.82055/2.08536/3.14133 
			& 0.83502/1.94564/3.11697 & 1.16865/2.53896/3.60636 & 1.10773/2.44406/3.16840
			& 0.99891/2.12089/3.05541 & 1.16113/2.68963/4.01089 & 0.92387/1.99285/2.99215 \\
			($\theta$) 
			& (6.283)/(0.000)/(0.000) & (1.257)/(1.257)/(0.628) & (0.000)/(0.000)/(0.000)
			& (6.283)/(6.283)/(0.000) & (0.628)/(5.655)/(5.655) & (3.142)/(5.027)/(1.885)
			& (2.513)/(0.000)/(0.000) & (4.398)/(4.398)/(4.398) & (3.142)/(3.142)/(3.142) \\
			I-RP-GFRFT\(_\mathbf{R_{pitch}}\)  
			& 1.06277/2.51669/3.77586 & \textbf{0.85390/2.02921/3.14565} & 0.80825/1.81142/2.69476 
			& \textbf{0.83230/1.94564/3.11697} & 1.15019/2.43295/3.51819 & 0.89103/2.18751/3.12475 
			& 0.65679/1.58964/2.52349 & 0.91378/2.15563/3.21209 & \textbf{0.62323/1.60560/2.55872} \\
            \multirow{3}{*}{($\alpha,\theta$)}
			& (0.900,0.000)($\sigma$=0.5) & (0.600,1.257)($\sigma$=0.5) & (0.400,0.628)($\sigma$=0.5)
			& (0.900,1.885)($\sigma$=0.5) & (0.500,1.885)($\sigma$=0.5) & (0.300,1.885)($\sigma$=0.5)
			& (0.800,6.283)($\sigma$=0.5) & (0.600,0.000)($\sigma$=0.5) & (0.700,2.513)($\sigma$=0.5) \\  
			& (0.800,5.027)($\sigma$=1.0) & (0.600,1.257)($\sigma$=1.0) & (0.500,0.628)($\sigma$=1.0)
			& (1.000,6.283)($\sigma$=1.0) & (0.400,1.885)($\sigma$=1.0) & (0.800,1.885)($\sigma$=1.0)
			& (0.800,6.283)($\sigma$=1.0) & (0.600,0.000)($\sigma$=1.0) & (0.700,2.513)($\sigma$=1.0) \\ 
			& (0.800,5.027)($\sigma$=1.5) & (0.600,1.257)($\sigma$=1.5) & (0.500,0.628)($\sigma$=1.5)
			& (1.000,6.283)($\sigma$=1.5) & (0.400,1.885)($\sigma$=1.5) & (1.000,2.513)($\sigma$=1.5)
			& (0.800,0.000)($\sigma$=1.5) & (0.700,6.283)($\sigma$=1.5) & (0.700,2.513)($\sigma$=1.5) \\ 
			I-RP-GFRFT\(_\mathbf{R_{roll}}\)  
			& 1.06277/2.41092/3.61885 & 1.10655/2.50253/3.53129 & 0.80023/2.01967/3.08957 
			& 0.83502/1.94564/3.11697 & \textbf{0.93759/2.25501/3.43771} & \textbf{0.63573}/1.78107/2.75387 
			& 0.65679/1.58964/2.52349 & \textbf{0.84205/2.06051/3.21209} & 0.88136/2.14633/3.02735 \\
            \multirow{3}{*}{($\alpha,\theta$)}
			& (0.900,0.000)($\sigma$=0.5) & (0.500,3.770)($\sigma$=0.5) & (0.300,5.027)($\sigma$=0.5)
			& (1.000,0.000)($\sigma$=0.5) & (0.700,3.142)($\sigma$=0.5) & (0.300,5.027)($\sigma$=0.5)
			& (0.800,0.000)($\sigma$=0.5) & (0.600,1.257)($\sigma$=0.5) & (0.100,5.655)($\sigma$=0.5) \\  
			& (0.700,6.283)($\sigma$=1.0) & (0.800,1.257)($\sigma$=1.0) & (0.300,5.027)($\sigma$=1.0)
			& (1.000,0.000)($\sigma$=1.0) & (0.700,3.142)($\sigma$=1.0) & (0.300,5.027)($\sigma$=1.0)
			& (0.800,0.000)($\sigma$=1.0) & (0.600,1.257)($\sigma$=1.0) & (0.600,6.283)($\sigma$=1.0) \\ 
			& (0.700,6.283)($\sigma$=1.5) & (0.900,1.257)($\sigma$=1.5) & (0.300,5.027)($\sigma$=1.5)
			& (1.000,0.000)($\sigma$=1.5) & (0.700,3.142)($\sigma$=1.5) & (0.800,3.770)($\sigma$=1.5)
			& (0.800,0.000)($\sigma$=1.5) & (0.700,0.000)($\sigma$=1.5) & (0.600,6.283)($\sigma$=1.5) \\ 
			I-RP-GFRFT\(_\mathbf{R_{yaw}}\)  
			& 1.06277/2.59296/3.99297 & 1.12119/2.42554/3.55226 & 0.80978/2.02104/2.99489 
			& 0.83502/1.94564/3.11697 & 1.08596/2.53896/3.60636 & 0.94066/2.11125/3.05474 
			& 0.65679/1.58964/2.52349 & 0.88764/2.15563/3.21209 & 0.92387/1.99285/2.99215 \\
            \multirow{3}{*}{($\alpha,\theta$)}
			& (0.900,0.000)($\sigma$=0.5) & (0.800,0.628)($\sigma$=0.5) & (0.800,3.142)($\sigma$=0.5)
			& (1.000,0.000)($\sigma$=0.5) & (0.500,5.027)($\sigma$=0.5) & (0.300,3.770)($\sigma$=0.5)
			& (0.800,0.000)($\sigma$=0.5) & (0.500,0.628)($\sigma$=0.5) & (1.000,3.142)($\sigma$=0.5) \\  
			& (0.900,0.000)($\sigma$=1.0) & (0.300,1.885)($\sigma$=1.0) & (0.800,1.885)($\sigma$=1.0)
			& (1.000,6.283)($\sigma$=1.0) & (1.000,5.655)($\sigma$=1.0) & (0.800,5.655)($\sigma$=1.0)
			& (0.800,6.283)($\sigma$=1.0) & (0.600,6.283)($\sigma$=1.0) & (1.000,3.142)($\sigma$=1.0) \\ 
			& (0.800,0.000)($\sigma$=1.5) & (0.400,1.885)($\sigma$=1.5) & (0.800,1.885)($\sigma$=1.5)
			& (1.000,0.000)($\sigma$=1.5) & (1.000,5.655)($\sigma$=1.5) & (0.800,5.655)($\sigma$=1.5)
			& (0.800,6.283)($\sigma$=1.5) & (0.700,6.283)($\sigma$=1.5) & (1.000,3.142)($\sigma$=1.5) \\ 
			II-RP-GFRFT\(_\mathbf{R_{pitch}}\)  
			& \textbf{1.02406/2.40531/3.57178} & 1.10776/2.70620/3.84711 & \textbf{0.71507/1.70623/2.62088} 
			& 0.83502/1.94564/3.11697 & 0.94405/2.36803/3.60381 & 0.73297/\textbf{1.67026/2.60500} 
			& 0.65679/1.58964/2.52349 & 0.91378/2.15563/3.21209 & 0.93884/2.19658/3.19570\\
            \multirow{3}{*}{($\alpha,\theta$)}
			& (0.900,3.142)($\sigma$=0.5) & (0.400,0.628)($\sigma$=0.5) & (0.500,3.770)($\sigma$=0.5)
			& (1.000,6.283)($\sigma$=0.5) & (0.800,3.770)($\sigma$=0.5) & (0.600,3.770)($\sigma$=0.5)
			& (0.800,0.000)($\sigma$=0.5) & (0.600,6.283)($\sigma$=0.5) & (0.500,3.142)($\sigma$=0.5) \\  
			& (0.400,3.770)($\sigma$=1.0) & (1.000,2.513)($\sigma$=1.0) & (0.500,3.770)($\sigma$=1.0)
			& (1.000,6.283)($\sigma$=1.0) & (0.800,3.770)($\sigma$=1.0) & (0.600,3.770)($\sigma$=1.0)
			& (0.800,6.283)($\sigma$=1.0) & (0.600,0.000)($\sigma$=1.0) & (0.600,3.142)($\sigma$=1.0) \\ 
			& (0.400,3.770)($\sigma$=1.5) & (0.900,1.257)($\sigma$=1.5) & (0.500,3.770)($\sigma$=1.5)
			& (1.000,0.000)($\sigma$=1.5) & (0.500,1.257)($\sigma$=1.5) & (0.600,3.770)($\sigma$=1.5)
			& (0.800,0.000)($\sigma$=1.5) & (0.700,6.283)($\sigma$=1.5) & (0.500,5.027)($\sigma$=1.5) \\ 
			II-RP-GFRFT\(_\mathbf{R_{roll}}\)  
			& 1.06277/2.59296/3.93963 & 0.86090/2.23903/3.50022 & 0.82055/2.08536/3.11060
			& 0.83502/1.94564/3.11697 & 1.03554/2.43761/3.58579 & 0.91084/1.86315/2.65751 
			& 0.65679/1.58964/2.52349 & 0.91378/2.15563/3.21209 & 0.88520/2.05553/3.03386 \\
            \multirow{3}{*}{($\alpha,\theta$)}
			& (0.900,0.000)($\sigma$=0.5) & (0.700,2.513)($\sigma$=0.5) & (1.000,0.000)($\sigma$=0.5)
			& (1.000,0.000)($\sigma$=0.5) & (1.000,1.257)($\sigma$=0.5) & (0.200,1.257)($\sigma$=0.5)
			& (0.800,0.000)($\sigma$=0.5) & (0.600,0.000)($\sigma$=0.5) & (0.200,3.142)($\sigma$=0.5) \\  
			& (0.900,0.000)($\sigma$=1.0) & (0.700,2.513)($\sigma$=1.0) & (1.000,0.000)($\sigma$=1.0)
			& (1.000,0.000)($\sigma$=1.0) & (0.300,0.628)($\sigma$=1.0) & (0.800,3.770)($\sigma$=1.0)
			& (0.800,0.000)($\sigma$=1.0) & (0.600,0.000)($\sigma$=1.0) & (0.300,5.655)($\sigma$=1.0) \\ 
			& (0.500,5.027)($\sigma$=1.5) & (0.700,2.513)($\sigma$=1.5) & (0.000,1.257)($\sigma$=1.5)
			& (1.000,0.000)($\sigma$=1.5) & (0.300,0.628)($\sigma$=1.5) & (0.800,3.770)($\sigma$=1.5)
			& (0.800,0.000)($\sigma$=1.5) & (0.700,0.000)($\sigma$=1.5) & (0.300,5.655)($\sigma$=1.5) \\ 
			II-RP-GFRFT\(_\mathbf{R_{yaw}}\)  
			& 1.06277/2.59296/3.99297 & 1.16299/2.75467/3.87883 & 0.82055/2.08536/3.14133
			& 0.83502/1.94564/3.11697 & 1.05085/2.50797/3.60636 & 1.04354/2.09684/2.97140 
			& 0.65679/1.58964/2.52349 & 0.91378/2.15563/3.21209 & 0.82654/1.98874/2.91978 \\
            \multirow{3}{*}{($\alpha,\theta$)}
			& (0.900,6.283)($\sigma$=0.5) & (0.400,2.513)($\sigma$=0.5) & (1.000,0.000)($\sigma$=0.5)
			& (1.000,0.000)($\sigma$=0.5) & (0.500,3.142)($\sigma$=0.5) & (0.700,5.655)($\sigma$=0.5)
			& (0.800,0.000)($\sigma$=0.5) & (0.600,6.283)($\sigma$=0.5) & (0.700,5.027)($\sigma$=0.5) \\  
			& (0.900,6.283)($\sigma$=1.0) & (1.000,1.257)($\sigma$=1.0) & (1.000,0.000)($\sigma$=1.0)
			& (1.000,0.000)($\sigma$=1.0) & (0.500,3.142)($\sigma$=1.0) & (0.700,5.655)($\sigma$=1.0)
			& (0.800,6.283)($\sigma$=1.0) & (0.600,6.283)($\sigma$=1.0) & (0.900,3.142)($\sigma$=1.0) \\ 
			& (0.800,0.000)($\sigma$=1.5) & (0.200,4.398)($\sigma$=1.5) & (1.000,0.000)($\sigma$=1.5)
			& (1.000,0.000)($\sigma$=1.5) & (1.000,5.655)($\sigma$=1.5) & (0.700,5.655)($\sigma$=1.5)
			& (0.800,0.000)($\sigma$=1.5) & (0.700,0.000)($\sigma$=1.5) & (0.900,3.142)($\sigma$=1.5) \\ 
			\bottomrule
		\end{tabular}
	}
\end{table}

\begin{table}[h!]
	\centering
	\caption{Denoising results on the COVID-19 dataset.}
	\label{tab:Covid_denoising_results}
	\resizebox{\textwidth}{!}{
		\scriptsize
		\begin{tabular}{@{}c*{9}{c}@{}}
			\toprule
			\multirow{4}{*}{Method} & \multicolumn{3}{c}{2-NN} & \multicolumn{3}{c}{5-NN} & \multicolumn{3}{c}{10-NN} \\
			\cmidrule(lr){2-4} \cmidrule(lr){5-7} \cmidrule(lr){8-10}
			& \multicolumn{1}{c}{t=100} & \multicolumn{1}{c}{t=200} & \multicolumn{1}{c}{t=300} 
			& \multicolumn{1}{c}{t=100} & \multicolumn{1}{c}{t=200} & \multicolumn{1}{c}{t=300} 
			& \multicolumn{1}{c}{t=100} & \multicolumn{1}{c}{t=200} & \multicolumn{1}{c}{t=300} \\
			\cmidrule(lr){2-4} \cmidrule(lr){5-7} \cmidrule(lr){8-10}
			& $\sigma$=0.5/1.0/1.5 & $\sigma$=0.5/1.0/1.5 & $\sigma$=0.5/1.0/1.5 
			& $\sigma$=0.5/1.0/1.5 & $\sigma$=0.5/1.0/1.5 & $\sigma$=0.5/1.0/1.5 
			& $\sigma$=0.5/1.0/1.5 & $\sigma$=0.5/1.0/1.5 & $\sigma$=0.5/1.0/1.5 \\
			\midrule
			GFRFT
			& 0.21605/0.56884/0.92413 & 0.25257/0.55586/0.73122 & 0.27779/0.59189/0.78180 
			& \textbf{0.18737/0.50713/0.84122} & 0.23814/0.47578/0.64260 & 0.26868/0.52915/0.69847
			& 0.17947/0.47275/0.76300 & 0.41169/0.66930/0.80413 & 0.45324/0.72299/0.86524 \\
			($\alpha$) 
			& (0.00)/(0.00)/(0.00) & (0.80)/(0.90)/(0.90) & (0.90)/(0.90)/(1.00)
			& (0.10)/(0.10)/(0.20) & (0.40)/(0.50)/(0.50) & (0.40)/(0.40)/(0.50)
			& (0.10)/(0.10)/(0.20) & (1.00)/(0.90)/(0.90) & (0.70)/(0.80)/(0.90) \\
			AGFT\(_\mathbf{R_{pitch}}\)  
			& 0.45909/0.93715/1.27036 & 0.30780/0.58125/0.72862 & 0.30991/0.60420/0.77251
			& 0.47972/0.87345/1.16260 & 0.30708/0.53004/0.67242 & 0.33008/0.54854/0.70014 
			& 0.51090/0.97704/1.27852 & 0.34258/0.55485/0.68853 & 0.36857/0.60106/0.74374 \\
			($\theta$) 
			& (0.000)/(0.000)/(6.283) & (6.283)/(4.398)/(4.398) & (6.283)/(6.283)/(4.398)
			& (2.513)/(0.000)/(0.000) & (3.142)/(4.398)/(4.398) & (4.398)/(4.398)/(4.398)
			& (3.770)/(3.770)/(4.398) & (4.398)/(5.027)/(5.027) & (4.398)/(5.027)/(5.027) \\
			AGFT\(_\mathbf{R_{roll}}\)  
			& 0.45909/0.93715/1.27036 & 0.30780/0.54193/0.69620 & 0.30991/0.55068/0.71966
			& 0.50615/0.87345/1.16260 & 0.35550/0.58028/0.72088 & 0.38714/0.61103/0.74828
			& 0.51674/0.97821/1.29539 & 0.36278/0.56754/0.70418 & 0.38596/0.61886/0.76508 \\
			($\theta$) 
			& (0.000)/(0.000)/(0.000) & (0.000)/(4.398)/(4.398) & (0.000)/(4.398)/(4.398)
			& (0.000)/(0.000)/(0.000) & (4.398)/(4.398)/(4.398) & (4.398)/(2.513)/(2.513)
			& (0.628)/(0.628)/(0.628) & (0.628)/(0.628)/(0.628) & (5.655)/(0.628)/(0.628) \\
			AGFT\(_\mathbf{R_{yaw}}\)  
			& 0.45909/0.93715/1.26588 & 0.30780/0.56902/0.70843 & 0.30558/0.59871/0.74854 
			& 0.49422/0.87345/1.16260 & 0.39335/0.63217/0.76245 & 0.40137/0.68254/0.82084
			& 0.54253/0.96582/1.26605 & 0.33302/0.54113/0.68778 & 0.36301/0.58752/0.74536 \\
			($\theta$) 
			& (0.000)/(0.000)/(5.655) & (0.000)/(4.398)/(4.398) & (3.142)/(4.398)/(4.398)
			& (1.257)/(0.000)/(0.000) & (5.027)/(4.398)/(4.398) & (3.142)/(4.398)/(4.398)
			& (3.770)/(3.770)/(3.770) & (4.398)/(4.398)/(4.398) & (4.398)/(4.398)/(4.398) \\
			I-RP-GFRFT\(_\mathbf{R_{pitch}}\)  
			& 0.20705/0.54848/0.87974 & \textbf{0.16546/0.39190/0.55578} & \textbf{0.17266/0.41080/0.58794}
			& 0.18737/0.50713/0.84122 & 0.17455/0.38630/0.56429 & 0.18016/0.40790/0.60154
			& 0.17947/0.47275/0.76300 & 0.27020/0.53852/0.68853 & 0.32367/0.60106/0.74374 \\
            \multirow{3}{*}{($\alpha,\theta$)}
			& (0.100,5.027)($\sigma$=0.5) & (0.200,2.513)($\sigma$=0.5) & (0.200,2.513)($\sigma$=0.5)
			& (0.100,0.000)($\sigma$=0.5) & (0.200,5.027)($\sigma$=0.5) & (0.300,5.027)($\sigma$=0.5)
			& (0.100,6.283)($\sigma$=0.5) & (0.300,1.885)($\sigma$=0.5) & (0.300,1.885)($\sigma$=0.5) \\  
			& (0.100,5.027)($\sigma$=1.0) & (0.300,2.513)($\sigma$=1.0) & (0.300,2.513)($\sigma$=1.0)
			& (0.100,0.000)($\sigma$=1.0) & (0.300,5.027)($\sigma$=1.0) & (0.300,5.027)($\sigma$=1.0)
			& (0.100,6.283)($\sigma$=1.0) & (0.400,1.885)($\sigma$=1.0) & (1.000,5.027)($\sigma$=1.0) \\ 
			& (0.100,5.027)($\sigma$=1.5) & (0.300,2.513)($\sigma$=1.5) & (0.300,2.513)($\sigma$=1.5)
			& (0.200,0.000)($\sigma$=1.5) & (0.300,5.027)($\sigma$=1.5) & (0.300,5.027)($\sigma$=1.5)
			& (0.200,0.000)($\sigma$=1.5) & (1.000,5.027)($\sigma$=1.5) & (1.000,5.027)($\sigma$=1.5) \\ 
			I-RP-GFRFT\(_\mathbf{R_{roll}}\)  
			& 0.19049/0.50243/0.82389 & 0.25257/0.54128/0.69620 & 0.27779/0.55068/0.71966 
			& 0.18737/0.50713/0.84122 & \textbf{0.13674/0.33077/0.47974} & \textbf{0.16151/0.37194/0.54012}
			& 0.17947/0.47275/0.76300 & \textbf{0.19428/0.40573/0.55055} & \textbf{0.20304/0.43961/0.59593} \\
            \multirow{3}{*}{($\alpha,\theta$)}
			& (0.100,5.655)($\sigma$=0.5) & (0.800,0.000)($\sigma$=0.5) & (0.900,0.000)($\sigma$=0.5)
			& (0.100,0.000)($\sigma$=0.5) & (0.200,1.257)($\sigma$=0.5) & (0.200,1.257)($\sigma$=0.5)
			& (0.100,0.000)($\sigma$=0.5) & (0.300,6.283)($\sigma$=0.5) & (0.300,6.283)($\sigma$=0.5) \\  
			& (0.100,5.027)($\sigma$=1.0) & (0.900,4.398)($\sigma$=1.0) & (1.000,4.398)($\sigma$=1.0)
			& (0.100,0.000)($\sigma$=1.0) & (0.300,1.257)($\sigma$=1.0) & (0.300,1.257)($\sigma$=1.0)
			& (0.100,6.283)($\sigma$=1.0) & (0.400,6.283)($\sigma$=1.0) & (0.300,6.283)($\sigma$=1.0) \\ 
			& (0.100,5.027)($\sigma$=1.5) & (1.000,4.398)($\sigma$=1.5) & (1.000,4.398)($\sigma$=1.5)
			& (0.200,0.000)($\sigma$=1.5) & (0.300,1.257)($\sigma$=1.5) & (0.300,1.257)($\sigma$=1.5)
			& (0.200,6.283)($\sigma$=1.5) & (0.400,6.283)($\sigma$=1.5) & (0.400,6.283)($\sigma$=1.5) \\ 
			I-RP-GFRFT\(_\mathbf{R_{yaw}}\)  
			& \textbf{0.18571/0.48139/0.78938} & 0.20058/0.44229/0.61702 & 0.22136/0.47304/0.65871
			& 0.18737/0.50713/0.84122 & 0.20687/0.44894/0.62434 & 0.23507/0.50091/0.69034 
			& 0.17947/0.47275/0.76300 & 0.20658/0.45079/0.64118 & 0.26149/0.54271/0.74124 \\
            \multirow{3}{*}{($\alpha,\theta$)}
			& (0.100,3.770)($\sigma$=0.5) & (0.300,5.027)($\sigma$=0.5) & (0.300,1.257)($\sigma$=0.5)
			& (0.100,0.000)($\sigma$=0.5) & (0.300,1.257)($\sigma$=0.5) & (0.300,1.257)($\sigma$=0.5)
			& (0.100,0.000)($\sigma$=0.5) & (0.300,5.655)($\sigma$=0.5) & (0.200,5.655)($\sigma$=0.5) \\  
			& (0.100,3.770)($\sigma$=1.0) & (0.400,5.027)($\sigma$=1.0) & (0.400,1.257)($\sigma$=1.0)
			& (0.100,6.283)($\sigma$=1.0) & (0.400,1.257)($\sigma$=1.0) & (0.400,1.257)($\sigma$=1.0)
			& (0.100,0.000)($\sigma$=1.0) & (0.300,5.655)($\sigma$=1.0) & (0.300,5.655)($\sigma$=1.0) \\ 
			& (0.200,3.770)($\sigma$=1.5) & (0.400,5.027)($\sigma$=1.5) & (0.400,1.257)($\sigma$=1.5)
			& (0.200,6.283)($\sigma$=1.5) & (0.400,1.257)($\sigma$=1.5) & (0.400,1.257)($\sigma$=1.5)
			& (0.200,0.000)($\sigma$=1.5) & (0.300,5.655)($\sigma$=1.5) & (0.900,4.398)($\sigma$=1.5) \\ 
			II-RP-GFRFT\(_\mathbf{R_{pitch}}\)  
			& 0.21605/0.56884/0.92413 & 0.23204/0.46422/0.61394 & 0.27779/0.57113/0.72671
			& 0.18737/0.50713/0.84122 & 0.23814/0.45759/0.59545 & 0.26868/0.52915/0.69847
			& 0.17947/0.47275/0.76300 & 0.27756/0.47049/0.61394 & 0.30791/0.58640/0.73832 \\
            \multirow{3}{*}{($\alpha,\theta$)}
			& (0.000,3.142)($\sigma$=0.5) & (0.200,1.257)($\sigma$=0.5) & (0.900,6.283)($\sigma$=0.5)
			& (0.100,6.283)($\sigma$=0.5) & (0.400,6.283)($\sigma$=0.5) & (0.400,0.000)($\sigma$=0.5)
			& (0.100,0.000)($\sigma$=0.5) & (0.000,1.257)($\sigma$=0.5) & (0.100,5.027)($\sigma$=0.5) \\  
			& (0.000,3.142)($\sigma$=1.0) & (0.100,1.257)($\sigma$=1.0) & (0.500,1.257)($\sigma$=1.0)
			& (0.100,6.283)($\sigma$=1.0) & (0.100,4.398)($\sigma$=1.0) & (0.400,0.000)($\sigma$=1.0)
			& (0.100,6.283)($\sigma$=1.0) & (0.000,4.398)($\sigma$=1.0) & (0.100,4.398)($\sigma$=1.0) \\ 
			& (0.000,3.142)($\sigma$=1.5) & (0.000,4.398)($\sigma$=1.5) & (0.500,2.513)($\sigma$=1.5)
			& (0.200,6.283)($\sigma$=1.5) & (0.100,4.398)($\sigma$=1.5) & (0.500,0.000)($\sigma$=1.5)
			& (0.200,6.283)($\sigma$=1.5) & (0.000,4.398)($\sigma$=1.5) & (0.100,4.398)($\sigma$=1.5) \\ 
			II-RP-GFRFT\(_\mathbf{R_{roll}}\)  
			& 0.21605/0.56884/0.92413 & 0.25257/0.50736/0.67152 & 0.27779/0.52422/0.69890 
			& 0.18737/0.50713/0.84122 & 0.23814/0.47578/0.64260 & 0.26868/0.52915/0.69847 
			& 0.17947/0.47275/0.76300 & 0.35551/0.56754/0.70418 & 0.37150/0.61886/0.76508 \\
            \multirow{3}{*}{($\alpha,\theta$)}
			& (0.000,0.000)($\sigma$=0.5) & (0.800,0.000)($\sigma$=0.5) & (0.900,0.000)($\sigma$=0.5)
			& (0.100,0.000)($\sigma$=0.5) & (0.400,0.000)($\sigma$=0.5) & (0.400,0.000)($\sigma$=0.5)
			& (0.100,0.000)($\sigma$=0.5) & (0.600,0.628)($\sigma$=0.5) & (0.900,5.655)($\sigma$=0.5) \\  
			& (0.000,0.000)($\sigma$=1.0) & (0.800,4.398)($\sigma$=1.0) & (0.900,4.398)($\sigma$=1.0)
			& (0.100,0.000)($\sigma$=1.0) & (0.500,0.000)($\sigma$=1.0) & (0.400,0.000)($\sigma$=1.0)
			& (0.100,0.000)($\sigma$=1.0) & (1.000,0.628)($\sigma$=1.0) & (1.000,0.628)($\sigma$=1.0) \\ 
			& (0.000,0.000)($\sigma$=1.5) & (0.900,4.398)($\sigma$=1.5) & (0.900,4.398)($\sigma$=1.5)
			& (0.200,0.000)($\sigma$=1.5) & (0.500,0.000)($\sigma$=1.5) & (0.500,0.000)($\sigma$=1.5)
			& (0.200,0.000)($\sigma$=1.5) & (1.000,0.628)($\sigma$=1.5) & (1.000,0.628)($\sigma$=1.5) \\ 
			II-RP-GFRFT\(_\mathbf{R_{yaw}}\)  
			& 0.21605/0.56884/0.92413 & 0.24004/0.49112/0.66363 & 0.25706/0.53847/0.72662 
			& 0.18737/0.50713/0.84122 & 0.23814/0.47578/0.64260 & 0.26868/0.52915/0.69847 
			& \textbf{0.17506/0.47275/0.76300} & 0.33302/0.54113/0.68778 & 0.32862/0.58752/0.74536 \\ 
            \multirow{3}{*}{($\alpha,\theta$)}
			& (0.000,0.000)($\sigma$=0.5) & (0.200,3.142)($\sigma$=0.5) & (0.200,3.142)($\sigma$=0.5)
			& (0.100,6.283)($\sigma$=0.5) & (0.400,0.000)($\sigma$=0.5) & (0.400,6.283)($\sigma$=0.5)
			& (0.100,3.142)($\sigma$=0.5) & (1.000,4.398)($\sigma$=0.5) & (0.200,1.257)($\sigma$=0.5) \\  
			& (0.000,0.000)($\sigma$=1.0) & (0.300,3.142)($\sigma$=1.0) & (0.300,3.142)($\sigma$=1.0)
			& (0.100,6.283)($\sigma$=1.0) & (0.500,6.283)($\sigma$=1.0) & (0.400,6.283)($\sigma$=1.0)
			& (0.100,0.000)($\sigma$=1.0) & (1.000,4.398)($\sigma$=1.0) & (1.000,4.398)($\sigma$=1.0) \\ 
			& (0.000,3.142)($\sigma$=1.5) & (0.400,3.142)($\sigma$=1.5) & (0.300,3.142)($\sigma$=1.5)
			& (0.200,6.283)($\sigma$=1.5) & (0.500,0.000)($\sigma$=1.5) & (0.500,0.000)($\sigma$=1.5)
			& (0.200,0.000)($\sigma$=1.5) & (1.000,4.398)($\sigma$=1.5) & (1.000,4.398)($\sigma$=1.5) \\ 
			\bottomrule
		\end{tabular}
	}
\end{table}

\begin{table}[htbp]
\centering
\caption{Comparison with classical and GNN-style methods for real-data denoising. Values are MSEs for $\sigma=0.5/1.0/1.5$.}
\label{tab:real_data_baseline_comparison}
\resizebox{\textwidth}{!}{%
\scriptsize
\begin{tabular}{@{}l*{9}{c}@{}}
\toprule
\multirow{2}{*}{Method} & \multicolumn{3}{c}{SST} & \multicolumn{3}{c}{PM-25} & \multicolumn{3}{c}{COVID-19} \\
\cmidrule(lr){2-4} \cmidrule(lr){5-7} \cmidrule(lr){8-10}
& $t=100$ & $t=200$ & $t=300$ & $t=100$ & $t=200$ & $t=300$ & $t=100$ & $t=200$ & $t=300$ \\
\midrule
\multicolumn{10}{c}{Classical methods} \\
\midrule
Moving average
& 10.534/12.479/18.549 & 8.3659/8.4320/12.198 & 9.4830/13.657/24.074
& 7.9484/9.4826/10.665 & 5.7799/7.3422/8.3861 & 7.1183/13.586/8.1360
& 2.6752/4.4255/10.266 & 1.0156/1.3714/6.1877 & 3.8445/8.0091/18.889 \\
Median filter
& 9.6735/13.562/18.443 & 8.2112/10.862/17.274 & 11.410/18.420/26.826
& 7.3304/13.201/13.381 & 6.1318/10.152/12.729 & 7.2176/13.059/18.356
& 2.9000/4.1754/8.6838 & 1.4155/4.6757/11.191 & 3.6365/13.508/27.391 \\
Gaussian smoothing
& 10.313/13.535/28.728 & 8.2034/10.760/23.886 & 9.8014/19.323/27.464
& 8.7629/13.324/15.351 & 6.9912/11.026/11.452 & 7.2502/17.246/12.316
& 3.9211/7.9190/21.719 & 2.1238/4.8701/17.989 & 5.7046/14.672/22.372 \\
Savitzky--Golay
& 11.380/14.525/40.160 & 8.6244/9.0639/24.281 & 10.027/20.826/26.244
& 8.4623/10.790/13.338 & 7.2895/10.015/8.7308 & 6.1438/11.509/10.797
& 4.1771/6.7465/32.927 & 1.2357/2.3296/15.491 & 4.7512/15.518/21.633 \\
Tikhonov 1st diff.
& 9.2767/10.784/18.990 & 7.4670/7.5714/13.024 & 8.1218/13.714/23.221
& 7.6099/9.1306/9.7108 & 5.7333/6.4385/6.7560 & 5.8301/10.856/7.0344
& 2.4293/4.2510/11.857 & 0.9591/1.3972/6.6473 & 3.6046/8.5370/18.154 \\
1D total variation
& 9.0159/9.7467/11.948 & 7.7512/7.6369/8.6138 & 7.5687/10.788/21.757
& 6.8790/7.7334/7.3157 & 5.4401/5.6094/6.2420 & 5.5235/10.020/5.5962
& 1.4028/1.9696/4.3384 & 0.4388/0.4718/1.2929 & 2.1405/5.1568/16.240 \\
FFT soft threshold
& 8.1711/14.514/26.190 & 8.0214/10.039/11.490 & 5.7594/12.535/46.869
& 6.1365/11.517/11.565 & 6.3780/8.8112/8.7445 & 4.8686/5.4773/5.5990
& 0.0023/0.0005/0.0005 & 0.2706/0.2756/0.2756 & 1.1262/1.1262/1.1262 \\
\midrule
\multicolumn{10}{c}{GNN-style methods} \\
\midrule
GCN
& 12.143/49.225/129.87 & 15.536/58.681/166.04 & 16.638/56.229/145.46
& 30.233/81.967/168.04 & 29.799/90.069/141.08 & 32.034/91.155/219.40
& 102.11/560.79/1088.2 & 112.49/487.68/1365.5 & 141.02/512.03/1103.3 \\
GraphSAGE
& 13.219/49.516/134.71 & 16.551/62.109/174.11 & 18.077/59.829/154.08
& 26.495/77.793/163.13 & 26.604/90.189/142.14 & 26.391/97.869/216.07
& 105.54/534.57/1061.6 & 113.41/476.17/1303.4 & 139.70/507.63/1139.1 \\
ChebNet
& 17.780/57.666/156.15 & 20.713/74.389/197.05 & 23.029/69.090/173.72
& 27.752/84.766/177.71 & 26.723/88.318/137.46 & 24.959/81.263/208.39
& 100.24/550.23/1075.4 & 111.21/470.84/1353.1 & 141.26/489.90/1095.5 \\
APPNP
& 14.433/50.481/136.69 & 16.690/62.912/171.99 & 19.061/60.141/151.56
& 27.337/86.969/165.93 & 27.839/91.821/137.87 & 29.966/88.134/215.57
& 108.42/541.40/1080.4 & 121.05/464.39/1350.0 & 152.01/498.72/1195.6 \\
GCNII
& 14.659/51.254/137.97 & 18.309/62.368/181.23 & 19.586/58.895/156.80
& 29.936/91.483/157.30 & 29.547/86.331/138.12 & 31.025/78.289/205.93
& 100.32/557.13/1083.1 & 111.87/487.75/1346.8 & 143.96/516.38/1128.0 \\
GAT
& 14.600/53.463/136.68 & 15.988/65.638/174.92 & 17.702/59.497/157.60
& 28.065/101.12/195.69 & 28.217/89.591/136.77 & 26.421/80.508/222.25
& 108.58/533.91/1125.8 & 113.20/472.29/1267.0 & 145.92/528.93/1101.8 \\
\midrule
\textbf{Ours}{\tiny (Best RP-GFRFT)}
& \textbf{0.5948/1.9991/3.8304}$^{\mathrm{A}}$ & \textbf{0.5069/1.6507/3.2597}$^{\mathrm{B}}$ & \textbf{0.6292/2.0586/3.9332}$^{\mathrm{A}}$
& \textbf{0.6568/1.5896/2.5235}$^{\mathrm{P}}$ & \textbf{0.8421/2.0292/3.1457}$^{\mathrm{Q}}$ & \textbf{0.6232/1.6056/2.5587}$^{\mathrm{P}}$
& \textbf{0.1751/0.4728/0.7630}$^{\mathrm{C}}$ & \textbf{0.1367/0.3308/0.4797}$^{\mathrm{D}}$ & \textbf{0.1615/0.3719/0.5401}$^{\mathrm{E}}$ \\
\bottomrule
\end{tabular}%
}
\vspace{3pt}

{\footnotesize \textbf{Note:} Each entry is an MSE triplet corresponding to $\sigma=0.5/1.0/1.5$, and bold entries denote the best triplet in each column. The last row reports the strongest RP-GFRFT result in each column, and the superscript indicates the corresponding configuration. \textbf{RP-GFRFT code:} $\mathrm{A}$=I-RP-GFRFT$_{\mathbf{R}_{\text{pitch}}}$ with $k=5/5/10$; $\mathrm{B}$=I-RP-GFRFT$_{\mathbf{R}_{\text{pitch}}}$ with $k=5/5/5$; $\mathrm{P}$=I-RP-GFRFT$_{\mathbf{R}_{\text{pitch}}}$ with $k=10/10/10$; $\mathrm{Q}$=I-RP-GFRFT with $\mathbf{R}_{\text{roll}}/\mathbf{R}_{\text{pitch}}/\mathbf{R}_{\text{pitch}}$ and $k=10/2/2$; $\mathrm{C}$=II-RP-GFRFT$_{\mathbf{R}_{\text{yaw}}}$ with $k=10/10/10$; $\mathrm{D}$=I-RP-GFRFT$_{\mathbf{R}_{\text{roll}}}$ with $k=5/5/5$ at $t=200$; $\mathrm{E}$=I-RP-GFRFT$_{\mathbf{R}_{\text{roll}}}$ with $k=5/5/5$ at $t=300$.}
\end{table}

\subsection{Image Denoising}
We further evaluate the proposed RP-GFRFT framework on the Set12 dataset, a widely recognized benchmark for image denoising tasks, which consists of 128×128 cropped regions of natural images including Peppers, Starfish, Monarch, and Parrot~\cite{imagedata}. Additive white Gaussian noise is injected into each image to simulate realistic noisy environments, with noise standard deviations configured as $\sigma=20$, $\sigma=30$, and $\sigma=40$. Each image is partitioned into non-overlapping 8×8 blocks. Each block is reshaped into a one-dimensional vector, which is further modeled as a graph signal on a 4-NN graph. The 4-NN graph is constructed based on the Euclidean distances between pixel coordinates to capture local spatial dependencies within the image block. Unlike low-dimensional temporal data, grid search is computationally infeasible for high-dimensional image data due to excessive computational overhead. Thus, the fractional order $\alpha$ and rotation angle $\theta$ are optimized via gradient descent to minimize the MSE, with a learning rate of 0.01 and a total of 1000 training epochs. All initial parameters are set to correspond to the GFT to ensure theoretical consistency at the onset of optimization. For qualitative visualization purposes, the best-performing rotation matrix (pitch, roll, or yaw) is selected for each method (AGFT, I-RP-GFRFT, and II-RP-GFRFT) to highlight their peak denoising capabilities.

\begin{table}[htbp]
	\centering
	\caption{Image denoising results on Set12 dataset.}
	\label{tab:denoising_results_set12}
	\resizebox{\textwidth}{!}{
		\scriptsize%
		\begin{tabular}{@{}c*{12}{c}@{}}
			\toprule
			\multirow{3}{*}{Method} & \multicolumn{3}{c}{Peppers} & \multicolumn{3}{c}{Starfish} & \multicolumn{3}{c}{Monarch} & \multicolumn{3}{c}{Parrot} \\
			\cmidrule(lr){2-4} \cmidrule(lr){5-7} \cmidrule(lr){8-10} \cmidrule(lr){11-13}
			& $\sigma$=20 & $\sigma$=30 & $\sigma$=40 & $\sigma$=20 & $\sigma$=30 & $\sigma$=40 & $\sigma$=20 & $\sigma$=30 & $\sigma$=40 & $\sigma$=20 & $\sigma$=30 & $\sigma$=40 \\
			\cmidrule(lr){2-4} \cmidrule(lr){5-7} \cmidrule(lr){8-10} \cmidrule(lr){11-13}
			& PSNR (MSE, SSIM) & PSNR (MSE, SSIM) & PSNR (MSE, SSIM) & PSNR (MSE, SSIM) & PSNR (MSE, SSIM) & PSNR (MSE, SSIM) & PSNR (MSE, SSIM) & PSNR (MSE, SSIM) & PSNR (MSE, SSIM) & PSNR (MSE, SSIM) & PSNR (MSE, SSIM) & PSNR (MSE, SSIM) \\
			\midrule
			GFRFT 
			& 46.449 (2.265e-05, 0.9915) & 44.321 (3.697e-05, 0.9846) & 42.217 (6.002e-05, 0.9805) 
			& 45.617 (2.744e-05, 0.9969) & 42.699 (5.371e-05, 0.9941) & 41.548 (7.002e-05, 0.9930) 
			& 45.624 (2.739e-05, 0.9950) & 44.007 (3.975e-05, 0.9934) & 41.648 (6.841e-05, 0.9908) 
			& 46.183 (2.408e-05, 0.9973) & 48.330 (1.469e-05, 0.9981) & 41.272 (7.462e-05, 0.9930) \\
			AGFT\(_\mathbf{R_{pitch}}\)
			& 38.661 (1.361e-04, 0.9550) & 35.210 (3.013e-04, 0.9127) & 32.912 (5.114e-04, 0.8697) 
			& 38.135 (1.536e-04, 0.9818) & 35.553 (2.784e-04, 0.9679) & 32.318 (5.864e-04, 0.9547) 
			& 38.854 (1.302e-04, 0.9736) & 35.536 (2.795e-04, 0.9526) & 33.803 (4.166e-04, 0.9369) 
			& 39.070 (1.239e-04, 0.9888) & 36.148 (2.428e-04, 0.9818) & 32.534 (5.579e-04, 0.9669) \\
			AGFT\(_\mathbf{R_{roll}}\) 
			& 38.362 (1.458e-04, 0.9487) & 35.757 (2.656e-04, 0.9105) & 32.898 (5.131e-04, 0.8585) 
			& 35.173 (3.039e-04, 0.9761) & 35.164 (3.045e-04, 0.9714) & 33.006 (5.005e-04, 0.9566) 
			& 38.419 (1.439e-04, 0.9718) & 36.190 (2.404e-04, 0.9544) & 34.973 (3.182e-04, 0.9388) 
			& 38.736 (1.338e-04, 0.9896) & 36.237 (2.379e-04, 0.9818) & 33.835 (4.135e-04, 0.9754) \\
			AGFT\(_\mathbf{R_{yaw}}\) 
			& 39.617 (1.092e-04, 0.9633) & 36.646 (2.165e-04, 0.9353) & 34.307 (3.710e-04, 0.9119) 
			& 40.976 (7.986e-05, 0.9923) & 37.539 (1.763e-04, 0.9795) & 34.700 (3.389e-04, 0.9658) 
			& 40.048 (9.891e-05, 0.9784) & 36.291 (2.349e-04, 0.9710) & 35.241 (2.992e-04, 0.9556) 
			& 37.826 (1.650e-04, 0.9876) & 36.731 (2.123e-04, 0.9840) & 34.351 (3.672e-04, 0.9783) \\
			I-RP-GFRFT\(_\mathbf{R_{pitch}}\)
			& 50.182 (9.589e-06, 0.9965) & 48.690 (1.352e-05, 0.9943) & 45.746 (2.663e-05, 0.9894) 
			& 50.789 (8.339e-06, 0.9990) & 47.739 (1.683e-05, 0.9982) & 45.393 (2.888e-05, 0.9972) 
			& 50.617 (8.676e-06, 0.9982) & 48.451 (1.429e-05, 0.9978) & 45.435 (2.861e-05, 0.9956) 
			& 51.908 (6.445e-06, 0.9993) & 47.767 (1.672e-05, 0.9979) & 44.970 (3.184e-05, 0.9977) \\
			I-RP-GFRFT\(_\mathbf{R_{roll}}\)
			& 51.467 (7.134e-06, 0.9972) & 48.678 (1.356e-05, 0.9944) & 45.953 (2.539e-05, 0.9918) 
			& 49.836 (1.038e-05, 0.9987) & 46.384 (2.299e-05, 0.9976) & 46.011 (2.506e-05, 0.9974) 
			& 51.430 (7.194e-06, 0.9982) & 48.372 (1.455e-05, 0.9969) & 45.386 (2.893e-05, 0.9950) 
			& 49.640 (1.086e-05, 0.9991) & 49.060 (1.242e-05, 0.9987) & 45.619 (2.742e-05, 0.9977) \\
			I-RP-GFRFT\(_\mathbf{R_{yaw}}\)
			& 52.362 (5.804e-06, 0.9979) & \textbf{50.182 (9.589e-06, 0.9963)} & 46.832 (2.074e-05, 0.9932) 
			& 51.656 (6.829e-06, 0.9993) & 48.465 (1.424e-05, \textbf{0.9984}) & 46.058 (2.479e-05, 0.9972) 
			& 51.761 (6.666e-06, 0.9989) & 48.132 (1.537e-05, \textbf{0.9982}) & \textbf{46.961 (2.013e-05, 0.9976)} 
			& 51.557 (6.987e-06, 0.9992) & 47.336 (1.847e-05, 0.9984) & 46.192 (2.403e-05, 0.9979) \\
			II-RP-GFRFT\(_\mathbf{R_{pitch}}\)
			& 50.490 (8.932e-06, 0.9968) & 47.893 (1.624e-05,0.9943) & 45.862 (2.593e-05, 0.9909) 
			& 50.129 (9.707e-06, 0.9988) & 48.312 (1.475e-05, 0.9983) & 45.176 (3.037e-05, 0.9964) 
			& 50.396 (9.129e-06, 0.9984) & \textbf{48.535} (\textbf{1.401e-05}, 0.9976) & 45.672 (2.709e-05, 0.9960) 
			& 51.838 (6.549e-06, \textbf{0.9994}) & 48.330 (1.469e-05, 0.9981) & 44.645 (3.431e-05, 0.9971) \\
			II-RP-GFRFT\(_\mathbf{R_{roll}}\)
			& 50.849 (8.224e-06, 0.9965) & 48.536 (1.401e-05, 0.9943) & \textbf{47.611 (1.733e-05, 0.9933)} 
			& 49.723 (1.066e-05, 0.9988) & 46.773 (2.102e-05, 0.9981) & 45.597 (2.756e-05, 0.9970) 
			& \textbf{51.907} (\textbf{6.446e-06}, 0.9985) & 48.321 (1.472e-05, 0.9972) & 46.007 (2.508e-05, 0.9958) 
			& 50.606 (8.697e-06, 0.9992) & 48.531 (1.402e-05, 0.9985) & 45.469 (2.838e-05, 0.9981) \\
			II-RP-GFRFT\(_\mathbf{R_{yaw}}\)
			& \textbf{53.326 (4.649e-06, 0.9980)} & 50.138 (9.687e-06, 0.9959) & 46.681 (2.147e-05, 0.9931) 
			& \textbf{52.475 (5.656e-06, 0.9994)} & \textbf{49.166 (1.212e-05, 0.9984)} & \textbf{47.546 (1.760e-05, 0.9980)} 
			& 51.516 (7.053e-06, \textbf{0.9991}) & 48.317 (1.473e-05, \textbf{0.9982}) & 46.897 (2.043e-05, 0.9970) 
			& \textbf{52.173} (\textbf{6.063e-06}, 0.9993) & \textbf{49.085 (1.234e-05, 0.9988)} & \textbf{46.342 (2.322e-05, 0.9980)} \\
			\bottomrule
		\end{tabular}
	}
\end{table}

\begin{table}[htbp]
	\centering
	\caption{Comparison with classical image denoising methods on Set12.}
	\label{tab:image_given_noisy_classical_comparison}
	{\setlength{\tabcolsep}{2pt}
	\renewcommand{\arraystretch}{1.05}
	\resizebox{\textwidth}{!}{%
		\tiny
		\begin{tabular}{@{}l*{12}{c}@{}}
			\toprule
			\multirow{3}{*}{Method} & \multicolumn{3}{c}{Peppers} & \multicolumn{3}{c}{Starfish} & \multicolumn{3}{c}{Monarch} & \multicolumn{3}{c}{Parrot} \\
			\cmidrule(lr){2-4} \cmidrule(lr){5-7} \cmidrule(lr){8-10} \cmidrule(lr){11-13}
			& $\sigma$=20 & $\sigma$=30 & $\sigma$=40 & $\sigma$=20 & $\sigma$=30 & $\sigma$=40 & $\sigma$=20 & $\sigma$=30 & $\sigma$=40 & $\sigma$=20 & $\sigma$=30 & $\sigma$=40 \\
			\cmidrule(lr){2-4} \cmidrule(lr){5-7} \cmidrule(lr){8-10} \cmidrule(lr){11-13}
			& PSNR (MSE, SSIM) & PSNR (MSE, SSIM) & PSNR (MSE, SSIM) & PSNR (MSE, SSIM) & PSNR (MSE, SSIM) & PSNR (MSE, SSIM) & PSNR (MSE, SSIM) & PSNR (MSE, SSIM) & PSNR (MSE, SSIM) & PSNR (MSE, SSIM) & PSNR (MSE, SSIM) & PSNR (MSE, SSIM) \\
			\midrule
			Gaussian filter
			& 28.1649 (1.526e-03, 0.7322) & 26.4907 (2.244e-03, 0.6767) & 25.3689 (2.905e-03, 0.6468)
			& 26.4814 (2.248e-03, 0.8096) & 24.5720 (3.490e-03, 0.7600) & 23.1180 (4.878e-03, 0.6920)
			& 26.3723 (2.306e-03, 0.7670) & 24.4931 (3.554e-03, 0.7319) & 23.1114 (4.885e-03, 0.6604)
			& 24.2544 (3.755e-03, 0.7994) & 22.5440 (5.567e-03, 0.7217) & 20.9796 (7.981e-03, 0.6545) \\
			Median filter
			& 27.2858 (1.868e-03, 0.6763) & 25.7008 (2.691e-03, 0.6759) & 24.5629 (3.497e-03, 0.5978)
			& 25.1082 (3.084e-03, 0.7800) & 23.3925 (4.579e-03, 0.6964) & 21.9028 (6.452e-03, 0.6220)
			& 25.5992 (2.755e-03, 0.7785) & 23.6307 (4.334e-03, 0.6801) & 21.9732 (6.349e-03, 0.6008)
			& 22.5533 (5.555e-03, 0.7278) & 21.3995 (7.245e-03, 0.6674) & 20.3158 (9.299e-03, 0.6148) \\
			Bilateral filter
			& 28.9923 (1.261e-03, 0.7688) & 26.5548 (2.211e-03, 0.6685) & 25.2850 (2.961e-03, 0.7087)
			& 26.6428 (2.166e-03, 0.8456) & 24.7096 (3.381e-03, 0.7728) & 23.2604 (4.720e-03, 0.7066)
			& 27.3544 (1.839e-03, 0.8443) & 25.0269 (3.143e-03, 0.7574) & 23.4387 (4.530e-03, 0.6849)
			& 25.5732 (2.771e-03, 0.8256) & 23.0800 (4.920e-03, 0.7475) & 21.2788 (7.449e-03, 0.6980) \\
			Non-local means (NLM)
			& 30.0142 (9.967e-04, 0.8485) & 27.9798 (1.592e-03, 0.7795) & 26.3599 (2.312e-03, 0.7119)
			& 26.8574 (2.062e-03, 0.8379) & 24.5906 (3.475e-03, 0.7739) & 22.7717 (5.282e-03, 0.6898)
			& 27.5878 (1.743e-03, 0.8859) & 25.3729 (2.902e-03, 0.8257) & 23.9741 (4.005e-03, 0.7603)
			& 25.8203 (2.618e-03, 0.8204) & 23.7166 (4.250e-03, 0.7579) & 22.2195 (5.999e-03, 0.7219) \\
			Wavelet shrinkage
			& 27.8859 (1.627e-03, 0.7982) & 26.1207 (2.443e-03, 0.7408) & 24.7925 (3.317e-03, 0.6950)
			& 25.1118 (3.082e-03, 0.7784) & 23.2486 (4.733e-03, 0.7138) & 22.0057 (6.301e-03, 0.6524)
			& 25.3943 (2.888e-03, 0.8272) & 23.5117 (4.455e-03, 0.7595) & 22.2718 (5.927e-03, 0.7099)
			& 23.5021 (4.465e-03, 0.7640) & 21.5069 (7.068e-03, 0.7082) & 20.2605 (9.418e-03, 0.6554) \\
			Total variation denoising
			& 22.1013 (6.164e-03, 0.4035) & 18.6269 (1.372e-02, 0.2826) & 16.2556 (2.368e-02, 0.2126)
			& 22.2733 (5.925e-03, 0.6462) & 18.9261 (1.281e-02, 0.5085) & 16.6193 (2.178e-02, 0.4096)
			& 22.1545 (6.089e-03, 0.5922) & 18.8140 (1.314e-02, 0.4721) & 16.5451 (2.216e-02, 0.3898)
			& 22.6475 (5.436e-03, 0.7101) & 19.4022 (1.148e-02, 0.6012) & 17.1550 (1.925e-02, 0.5187) \\
			DCT soft-thresholding
			& 27.4271 (1.808e-03, 0.7314) & 25.3888 (2.891e-03, 0.6194) & 23.9682 (4.010e-03, 0.6205)
			& 25.1537 (3.052e-03, 0.7615) & 22.7253 (5.339e-03, 0.6516) & 21.1033 (7.757e-03, 0.5739)
			& 25.4752 (2.835e-03, 0.7513) & 22.9552 (5.064e-03, 0.6371) & 21.3506 (7.327e-03, 0.6059)
			& 24.6199 (3.451e-03, 0.7836) & 22.0555 (6.229e-03, 0.7043) & 20.3103 (9.311e-03, 0.6319) \\
			Anisotropic diffusion
			& 27.1284 (1.937e-03, 0.7312) & 21.2760 (7.454e-03, 0.3883) & 17.4951 (1.780e-02, 0.2456)
			& 25.0236 (3.145e-03, 0.7980) & 20.8695 (8.186e-03, 0.6065) & 17.6835 (1.705e-02, 0.4556)
			& 25.7604 (2.654e-03, 0.7977) & 21.0783 (7.801e-03, 0.5679) & 17.7004 (1.698e-02, 0.4305)
			& 24.4844 (3.561e-03, 0.8000) & 20.9034 (8.122e-03, 0.6669) & 18.0571 (1.564e-02, 0.5488) \\
			\midrule
			\textbf{Ours}{\tiny (Best RP-GFRFT)}
			& \textbf{53.326 (4.649e-06, 0.9980)}$^{\mathrm{IIY}}$ & \textbf{50.182 (9.589e-06, 0.9963)}$^{\mathrm{IY}}$ & \textbf{47.611 (1.733e-05, 0.9933)}$^{\mathrm{IIR}}$
			& \textbf{52.475 (5.656e-06, 0.9994)}$^{\mathrm{IIY}}$ & \textbf{49.166 (1.212e-05, 0.9984)}$^{\mathrm{IIY}}$ & \textbf{47.546 (1.760e-05, 0.9980)}$^{\mathrm{IIY}}$
			& \textbf{51.907 (6.446e-06, 0.9985)}$^{\mathrm{IIR}}$ & \textbf{48.535 (1.401e-05, 0.9976)}$^{\mathrm{IIP}}$ & \textbf{46.961 (2.013e-05, 0.9976)}$^{\mathrm{IY}}$
			& \textbf{52.173 (6.063e-06, 0.9993)}$^{\mathrm{IIY}}$ & \textbf{49.085 (1.234e-05, 0.9988)}$^{\mathrm{IIY}}$ & \textbf{46.342 (2.322e-05, 0.9980)}$^{\mathrm{IIY}}$ \\
			\bottomrule
		\end{tabular}
	}}
	\vspace{3pt}

	{\footnotesize
	\textbf{Note:} Bold entries denote the best result in each column. ``Ours'' reports the best RP-GFRFT variant in Table~\ref{tab:denoising_results_set12}. \textbf{RP-GFRFT code:} $\mathrm{IY}$=I-RP-GFRFT$_{\mathbf{R}_{\text{yaw}}}$, $\mathrm{IIP}$=II-RP-GFRFT$_{\mathbf{R}_{\text{pitch}}}$, $\mathrm{IIR}$=II-RP-GFRFT$_{\mathbf{R}_{\text{roll}}}$, and $\mathrm{IIY}$=II-RP-GFRFT$_{\mathbf{R}_{\text{yaw}}}$.}
\end{table}
\begin{figure}[pos=t]
\centering
\includegraphics[width=\textwidth]{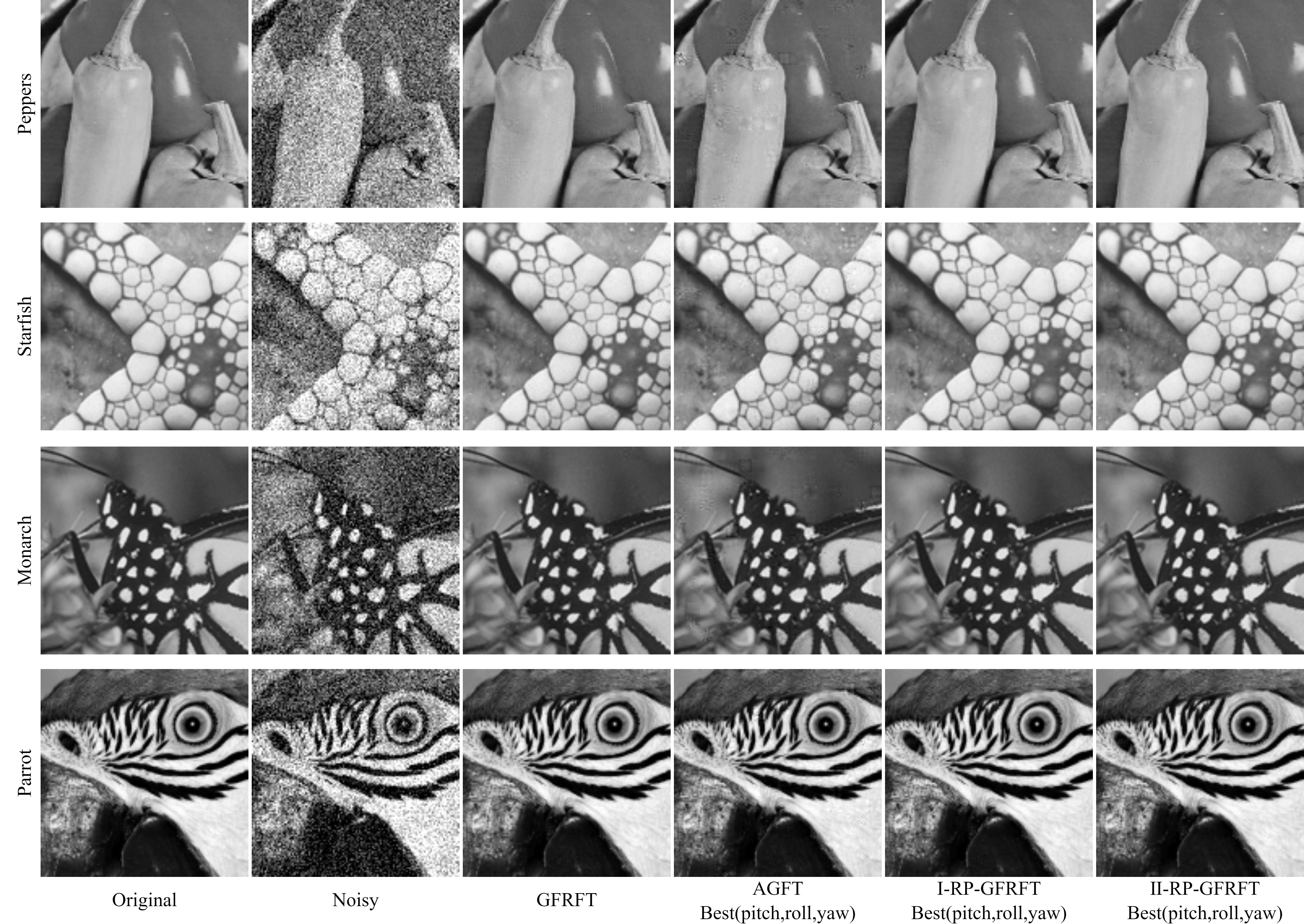}
\caption{Denoising results on Peppers, Starfish, Monarch, and Parrot images (\(\sigma=40\)). Columns: Original, Noisy, GFRFT, AGFT, I-RP-GFRFT, II-RP-GFRFT.}
\label{fig:image_denoising_results}
\end{figure}

\begin{figure}[pos=t]
\centering
\includegraphics[width=\textwidth]{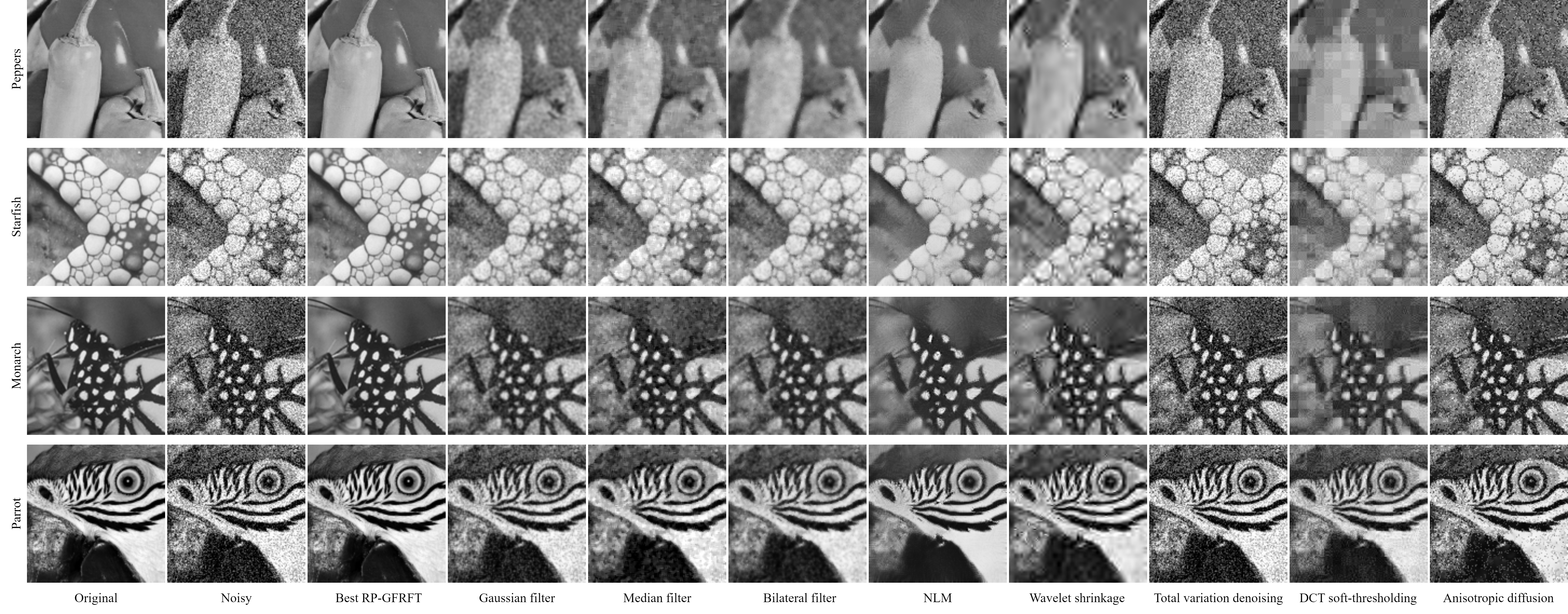}
\caption{Visual comparison with representative image denoising baselines (\(\sigma=40\)).}
\label{fig:image_baseline_visual_comparison}
\end{figure}

Table~\ref{tab:denoising_results_set12} summarizes the quantitative denoising performance (including MSE, PSNR, and SSIM) of all methods on the 128×128 cropped regions of the Set12 dataset. Both I-RP-GFRFT and II-RP-GFRFT variants consistently outperform GFRFT and AGFT (equipped with pitch, roll, or yaw rotation matrices) across all images and noise intensities, which benefits from the joint optimization of fractional order and rotation angle as well as the theoretical consistency guaranteed by initializing all parameters to correspond to GFT. For instance, on the Peppers image with $\sigma=20$ (moderate noise), II-RP-GFRFT with yaw rotation achieves the highest PSNR of 53.326 dB, the lowest MSE of 4.649e-06, and an SSIM of 0.9980, significantly exceeding GFRFT (46.449 dB, 2.265e-05, 0.9915) and the best-performing AGFT variant (AGFT with yaw rotation: 39.617 dB, 1.092e-04, 0.9633). Even at a higher noise level ($\sigma=40$), II-RP-GFRFT with roll rotation maintains superior performance on Peppers, achieving a PSNR of 47.611 dB compared to 42.217 dB of GFRFT, demonstrating its ability to suppress strong noise without over-smoothing while preserving structural details. The qualitative denoising results presented in Figure~\ref{fig:image_denoising_results} further validate these findings, where GFRFT fails to fully eliminate higher noise level ($\sigma=40$) in texture-rich regions (e.g., Peppers’ surfaces) and AGFT introduces noticeable blurring due to its defective degeneracy. In contrast, both I-RP-GFRFT and II-RP-GFRFT retain fine image details such as Starfish spines and Monarch wing patterns while effectively removing noise, underscoring RP-GFRFT’s practical effectiveness derived from a rigorous theoretical framework.

To further assess practical competitiveness, we compare the best RP-GFRFT results with representative classical image denoising methods, as reported in Table~\ref{tab:image_given_noisy_classical_comparison}. The selected methods include Gaussian filtering, median filtering~\cite{huang1979median}, bilateral filtering~\cite{tomasi1998bilateral}, non-local means (NLM)~\cite{buades2005nlm}, wavelet shrinkage~\cite{donoho1995}, total variation denoising~\cite{rudin1992}, DCT soft-thresholding~\cite{ahmed1974dct,donoho1995}, and anisotropic diffusion~\cite{perona1990anisotropic}, covering widely used spatial-, transform-, variational-, and diffusion-based denoising strategies. All classical image baselines are applied to the same noisy cropped images and evaluated using the same PSNR, MSE, and SSIM criteria; their filter sizes, thresholds, regularization weights, and iteration-related parameters are selected from fixed candidate settings based on the denoised MSE. A visual comparison with these representative baselines is shown in Figure~\ref{fig:image_baseline_visual_comparison}.

The comparison shows that these classical methods achieve reasonable denoising effects in some cases, but their performance is highly dependent on the image content and noise intensity. Spatial filtering methods tend to suppress noise at the cost of blurred edges and textures, whereas transform-domain thresholding methods preserve some structural information but remain less effective under stronger noise. NLM and anisotropic diffusion achieve competitive results in certain cases, yet they do not consistently provide the best PSNR, MSE, and SSIM across all test images. In contrast, RP-GFRFT obtains the best results in every column of Table~\ref{tab:image_given_noisy_classical_comparison}. This demonstrates that the proposed rotation-parameterized fractional spectral representation can better balance noise suppression and structural preservation for image graph signals than fixed spatial or transform-domain denoising rules.

\subsection{Point Cloud Denoising}
To validate RP-GFRFT’s effectiveness for high-dimensional geometric data, we evaluate its denoising performance on 3D point clouds using the Microsoft Voxelized dataset~\cite{pcdata}, consistent with cross-task experimental protocols. Additive Gaussian noise with standard deviations \(\sigma=20\), \(\sigma=30\), and \(\sigma=40\) is injected into clean point clouds to simulate low-, medium-, and high-noise environments. Each point cloud is partitioned into local patches (up to 100 vertices per patch). For each patch, 10-nearest neighbor (10-NN) graphs are constructed to model geometric dependencies: neighbors are identified via a k-d tree, edges are weighted using distance-based Gaussian kernels to emphasize proximity. Due to the high dimensionality and irregular topology of point cloud data, grid search is computationally infeasible. Instead, the fractional order \(\alpha\) and rotation angle \(\theta\) are optimized via gradient descent (learning rate = 0.01, 1000 epochs) to minimize MSE, with initial parameters set to correspond to GFT for theoretical consistency.

\begin{table}[htbp]
	\centering
	\caption{Point cloud denoising results on Microsoft Voxelized dataset.}
	\label{tab:denoising_results_Point}
	\resizebox{\textwidth}{!}{
		\scriptsize
		\begin{tabular}{@{}c*{12}{c}@{}}
			\toprule
			\multirow{3}{*}{Method} & \multicolumn{3}{c}{Sarah9} & \multicolumn{3}{c}{Ricardo9} & \multicolumn{3}{c}{Andrew9} & \multicolumn{3}{c}{David9} \\
			\cmidrule(lr){2-4} \cmidrule(lr){5-7} \cmidrule(lr){8-10} \cmidrule(lr){11-13}
			& $\sigma$=20 & $\sigma$=30 & $\sigma$=40 & $\sigma$=20 & $\sigma$=30 & $\sigma$=40 & $\sigma$=20 & $\sigma$=30 & $\sigma$=40 & $\sigma$=20 & $\sigma$=30 & $\sigma$=40 \\
			\cmidrule(lr){2-4} \cmidrule(lr){5-7} \cmidrule(lr){8-10} \cmidrule(lr){11-13}
			& PSNR (MSE) & PSNR (MSE) & PSNR (MSE) & PSNR (MSE) & PSNR (MSE) & PSNR (MSE) & PSNR (MSE) & PSNR (MSE) & PSNR (MSE) & PSNR (MSE) & PSNR (MSE) & PSNR (MSE) \\
			\midrule
			GFRFT 
			& 41.727 (6.719e-05) & 38.568 (1.391e-04) & 36.415 (2.283e-04) 
			& 38.200 (1.513e-04) & 36.244 (2.375e-04) & 34.936 (3.209e-04) 
			& 37.580 (1.746e-04) & 34.402 (3.629e-04) & 32.461 (5.674e-04) 
			& 37.749 (1.679e-04) & 35.166 (3.043e-04) & 33.325 (4.650e-04) \\
			AGFT\(_\mathbf{R_{pitch}}\)
			& 41.101 (7.761e-05) & 38.067 (1.561e-04) & 35.520 (2.805e-04) 
			& 37.415 (1.813e-04) & 35.577 (2.769e-04) & 34.240 (3.767e-04) 
			& 36.472 (2.253e-04) & 33.602 (4.363e-04) & 31.463 (7.140e-04) 
			& 36.545 (2.216e-04) & 34.192 (3.809e-04) & 32.706 (5.363e-04) \\
			AGFT\(_\mathbf{R_{roll}}\) 
			& 40.983 (7.975e-05) & 37.786 (1.665e-04) & 36.110 (2.449e-04) 
			& 37.604 (1.736e-04) & 35.781 (2.642e-04) & 34.652 (3.426e-04) 
			& 36.111 (2.448e-04) & 33.604 (4.361e-04) & 31.751 (6.682e-04) 
			& 36.719 (2.129e-04) & 34.611 (3.458e-04) & 32.906 (5.122e-04) \\
			AGFT\(_\mathbf{R_{yaw}}\) 
			& 40.934 (8.065e-05) & 37.804 (1.658e-04) & 35.565 (2.776e-04) 
			& 37.480 (1.786e-04) & 35.638 (2.730e-04) & 34.210 (3.793e-04) 
			& 36.454 (2.263e-04) & 33.789 (4.179e-04) & 31.494 (7.090e-04) 
			& 36.626 (2.175e-04) & 34.234 (3.772e-04) & 32.602 (5.492e-04) \\
			I-RP-GFRFT\(_\mathbf{R_{pitch}}\)
			& 46.455 (2.262e-05) & 43.480 (4.488e-05) & 41.283 (7.442e-05) 
			& 42.442 (5.699e-05) & 40.416 (9.086e-05) & 39.488 (1.125e-04) 
			& 42.789 (5.262e-05) & 39.519 (1.117e-04) & 37.308 (1.859e-04) 
			& 42.151 (6.094e-05) & 39.877 (1.029e-04) & 37.929 (1.611e-04) \\
			I-RP-GFRFT\(_\mathbf{R_{roll}}\)
			& 46.043 (2.487e-05) & 42.826 (5.217e-05) & 40.834 (8.253e-05) 
			& 42.037 (6.256e-05) & 40.463 (8.989e-05) & 39.222 (1.196e-04) 
			& 42.453 (5.685e-05) & 39.150 (1.216e-04) & 36.790 (2.094e-04) 
			& 42.090 (6.180e-05) & 39.728 (1.065e-04) & 37.885 (1.627e-04) \\
			I-RP-GFRFT\(_\mathbf{R_{yaw}}\)
			& \textbf{46.721 (2.127e-05)} & 43.764 (4.204e-05) & 41.247 (7.503e-05) 
			& 42.247 (5.960e-05) & 40.484 (8.945e-05) & 39.424 (1.142e-04) 
			& 42.666 (5.412e-05) & 39.517 (1.118e-04) & 37.076 (1.961e-04) 
			& 42.448 (5.691e-05) & 39.986 (1.003e-04) & 38.175 (1.522e-04) \\
			II-RP-GFRFT\(_\mathbf{R_{pitch}}\)
			& 46.436 (2.272e-05) & 43.625 (4.340e-05) & \textbf{41.392 (7.258e-05)} 
			& \textbf{42.540 (5.571e-05)} & 40.548 (8.814e-05) & 39.553 (1.109e-04) 
			& \textbf{42.841 (5.198e-05)} & \textbf{39.620 (1.091e-04)} & \textbf{37.491 (1.782e-04)} 
			& 42.278 (5.919e-05) & 39.997 (1.001e-04) & 38.275 (1.488e-04) \\
			II-RP-GFRFT\(_\mathbf{R_{roll}}\)
			& 46.110 (2.449e-05) & 43.018 (4.991e-05) & 41.374 (7.288e-05) 
			& 41.979 (6.340e-05) & 40.542 (8.826e-05) & 39.377 (1.154e-04) 
			& 42.524 (5.592e-05) & 39.175 (1.209e-04) & 36.979 (2.005e-04) 
			& 42.174 (6.061e-05) & 39.896 (1.024e-04) & 38.042 (1.570e-04) \\
			II-RP-GFRFT\(_\mathbf{R_{yaw}}\)
			& 46.710 (2.133e-05) & \textbf{43.911 (4.064e-05)} & 41.354 (7.321e-05) 
			& 42.309 (5.876e-05) & \textbf{40.624 (8.662e-05)} & \textbf{39.619 (1.092e-04)} 
			& 42.727 (5.337e-05) & 39.392 (1.150e-04) & 37.238 (1.889e-04) 
			& \textbf{42.475 (5.656e-05)} & \textbf{40.297 (9.338e-05)} & \textbf{38.454 (1.428e-04)} \\
			\bottomrule
		\end{tabular}
	}
\end{table}
\begin{figure}[pos=t]
\centering
\includegraphics[width=\textwidth]{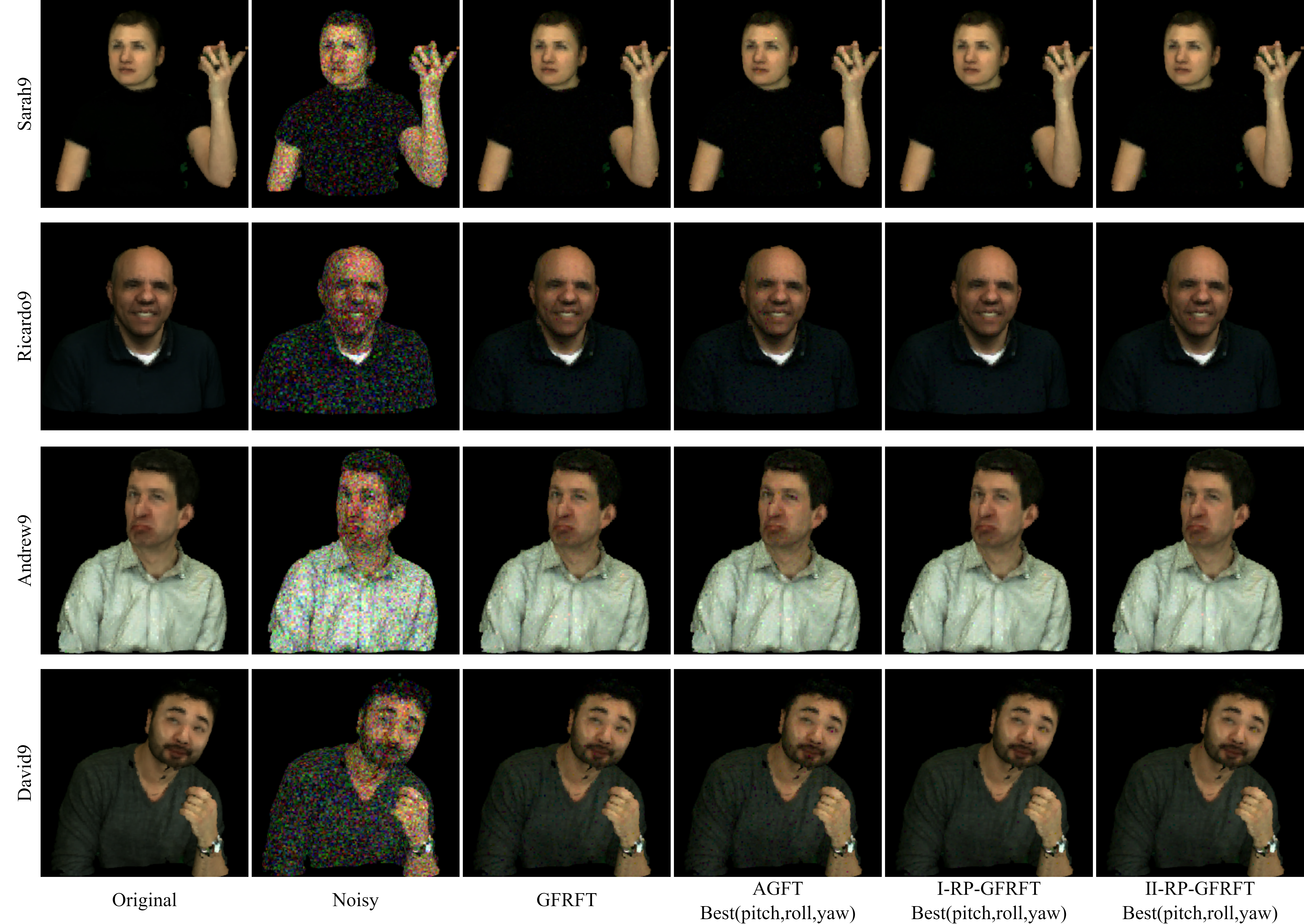}
\caption{Denoising results on \textit{Sarah9}, \textit{Ricardo9}, \textit{Andrew9}, and \textit{David9} (\(\sigma=40\)). Columns: Original, Noisy, GFRFT, AGFT, I-RP-GFRFT, II-RP-GFRFT.}
\label{fig:Point_denoising_results}
\end{figure}

\begin{figure}[pos=t]
\centering
\includegraphics[width=\textwidth]{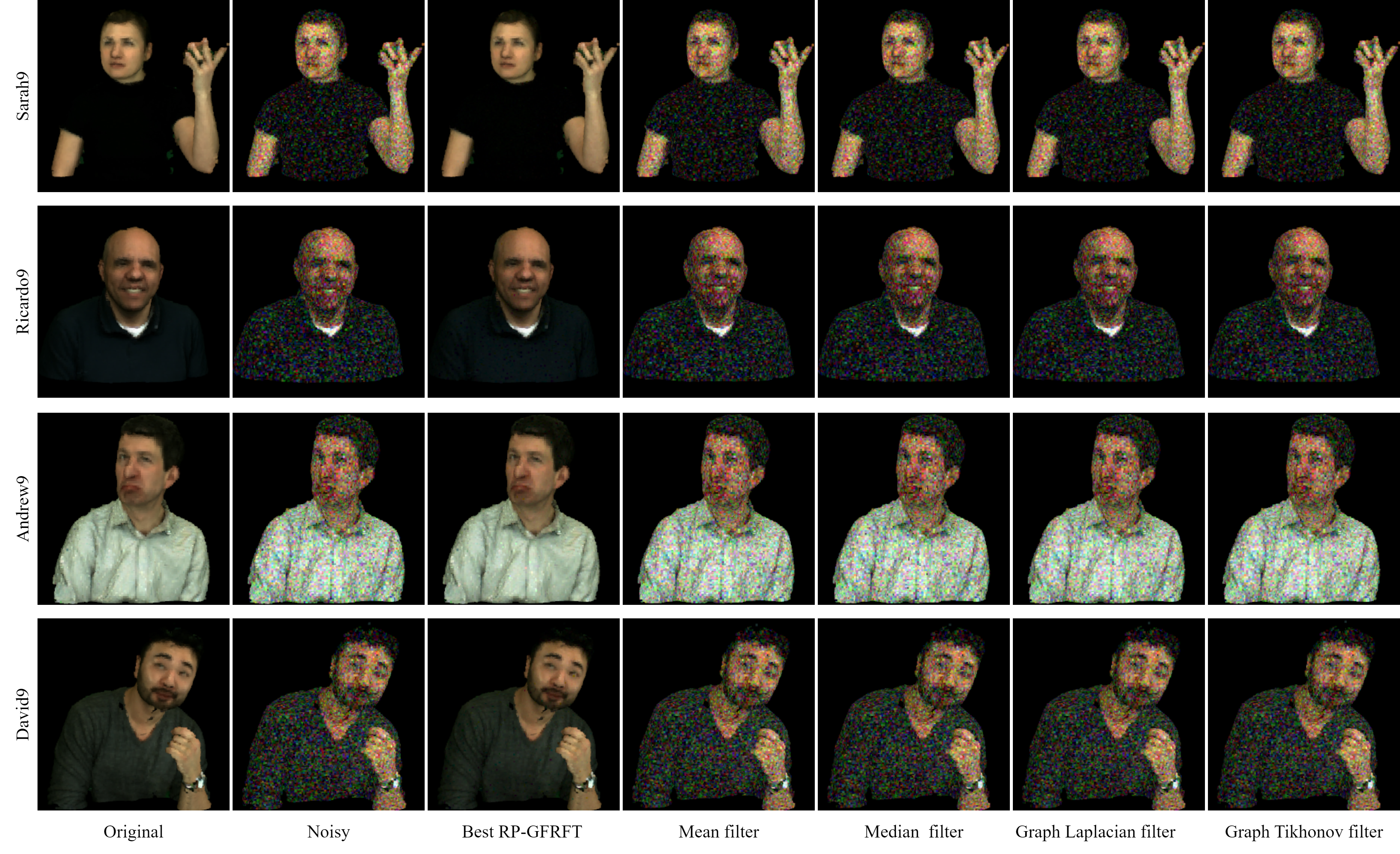}
\caption{Visual comparison with representative point-cloud graph-signal denoising baselines (\(\sigma=40\)).}
\label{fig:point_cloud_baseline_visual_comparison}
\end{figure}

Table~\ref{tab:denoising_results_Point} presents the quantitative denoising performance across all point cloud datasets. Both I-RP-GFRFT and II-RP-GFRFT (under pitch, roll, and yaw rotations) consistently outperform GFRFT and AGFT under all tested noise levels. In high-noise conditions (\(\sigma = 40\)), II-RP-GFRFT$_{\mathbf{R}_{\text{yaw}}}$ achieves the highest performance on \textit{David9}, yielding a PSNR of 38.454~dB and an MSE of 1.428e-04. This represents a 5.13~dB improvement in PSNR and a 69.3\% reduction in MSE compared with GFRFT (33.325~dB, 4.650e-04), and surpasses the best AGFT variant (AGFT$_{\mathbf{R}_{\text{roll}}}$, 32.906~dB, 5.122e-04) by 5.55~dB and 72.1\%, respectively. Similarly, on \textit{Andrew9}, II-RP-GFRFT$_{\mathbf{R}_{\text{pitch}}}$ attains a PSNR of 37.491~dB and an MSE of 1.782e-04, outperforming GFRFT (32.461~dB, 5.674e-04) by 5.03~dB in PSNR and 68.6\% in MSE reduction. 
Under moderate noise conditions (\(\sigma = 20\)), RP-GFRFT remains superior, demonstrating robust denoising capability even in less noisy environments. For \textit{Sarah9}, I-RP-GFRFT$_{\mathbf{R}_{\text{yaw}}}$ achieves a PSNR of 46.721~dB and an MSE of 2.127e-05, improving upon GFRFT (41.727~dB, 6.719e-05) by 4.99~dB in PSNR and 68.3\% in MSE reduction. Likewise, \textit{Ricardo9} reaches its optimal performance with II-RP-GFRFT$_{\mathbf{R}_{\text{pitch}}}$ (PSNR = 42.540~dB, MSE = 5.571e-05), outperforming GFRFT (38.200~dB, 1.513e-04) by 4.34~dB in PSNR and 63.2\% in MSE reduction. Figure~\ref{fig:Point_denoising_results} shows that RP-GFRFT can effectively preserve the subtle features of point cloud signal values while suppressing noise, which is consistent with the lower MSE and higher PSNR observed in the quantitative results. This performance benefits from the joint optimization of the fractional order $\alpha$ and rotation angle $\theta$, which enables better matching with the spectral structure of point cloud signals and provides valuable support for the denoising of 3D point cloud signals.

To further evaluate the proposed method against conventional graph-signal denoising strategies on point clouds, we compare the best RP-GFRFT results with representative baselines in Table~\ref{tab:point_cloud_classical_baseline_comparison}. The selected methods include Mean filter and Median filter as local neighborhood baselines, together with Graph Laplacian filter and Graph Tikhonov filter~\cite{tikhonov1977,bg00,bg03} as graph-regularized smoothing baselines on fixed point-cloud vertices. All baselines operate on the same point-cloud graph and the same noisy vertex signal values, without changing point locations or removing vertices; smoothing strengths and regularization weights are chosen from predefined candidate settings according to the denoised MSE. A visual comparison with these representative baselines is shown in Figure~\ref{fig:point_cloud_baseline_visual_comparison}.

The results indicate that conventional graph-signal smoothing methods provide limited recovery quality on the tested point-cloud signals. The Mean and median filters only aggregate neighboring signal values and therefore tend to oversmooth local variations. Graph Laplacian filter improves the results to some extent, while Graph Tikhonov filter gives the strongest performance among the listed conventional baselines. However, these methods remain far below RP-GFRFT across all datasets and noise levels. The consistent advantage of RP-GFRFT suggests that the rotation-parameterized fractional spectral representation is more effective for preserving point-cloud signal values while suppressing noise on the fixed graph structure.

\begin{table}[htbp]
\centering
\caption{Comparison with classical point-cloud graph-signal denoising methods on Microsoft Voxelized dataset.}
\label{tab:point_cloud_classical_baseline_comparison}
\resizebox{\textwidth}{!}{%
\scriptsize
\begin{tabular}{@{}l*{12}{c}@{}}
\toprule
\multirow{3}{*}{Method} & \multicolumn{3}{c}{Sarah9} & \multicolumn{3}{c}{Ricardo9} & \multicolumn{3}{c}{Andrew9} & \multicolumn{3}{c}{David} \\
\cmidrule(lr){2-4} \cmidrule(lr){5-7} \cmidrule(lr){8-10} \cmidrule(lr){11-13}
& $\sigma=20$ & $\sigma=30$ & $\sigma=40$ & $\sigma=20$ & $\sigma=30$ & $\sigma=40$ & $\sigma=20$ & $\sigma=30$ & $\sigma=40$ & $\sigma=20$ & $\sigma=30$ & $\sigma=40$ \\
\cmidrule(lr){2-4} \cmidrule(lr){5-7} \cmidrule(lr){8-10} \cmidrule(lr){11-13}
& \multicolumn{12}{c}{PSNR (MSE)} \\
\midrule
Mean filter
& 8.729 (1.340e-01) & 8.729 (1.340e-01) & 8.729 (1.340e-01)
& 9.538 (1.112e-01) & 9.538 (1.112e-01) & 9.538 (1.112e-01)
& 8.934 (1.278e-01) & 8.934 (1.278e-01) & 8.934 (1.278e-01)
& 8.241 (1.499e-01) & 8.241 (1.499e-01) & 8.241 (1.499e-01) \\
Median filter
& 8.393 (1.448e-01) & 8.393 (1.448e-01) & 8.393 (1.448e-01)
& 9.405 (1.147e-01) & 9.405 (1.147e-01) & 9.405 (1.147e-01)
& 8.739 (1.337e-01) & 8.739 (1.337e-01) & 8.739 (1.337e-01)
& 7.915 (1.616e-01) & 7.915 (1.616e-01) & 7.915 (1.616e-01) \\
Graph Laplacian filter
& 16.687 (2.144e-02) & 16.687 (2.144e-02) & 16.687 (2.144e-02)
& 17.496 (1.780e-02) & 17.496 (1.780e-02) & 17.496 (1.780e-02)
& 16.893 (2.045e-02) & 16.893 (2.045e-02) & 16.893 (2.045e-02)
& 16.200 (2.399e-02) & 16.200 (2.399e-02) & 16.200 (2.399e-02) \\
Graph Tikhonov filter
& 19.817 (1.043e-02) & 19.817 (1.043e-02) & 19.817 (1.043e-02)
& 19.991 (1.002e-02) & 19.991 (1.002e-02) & 19.991 (1.002e-02)
& 19.508 (1.120e-02) & 19.508 (1.120e-02) & 19.508 (1.120e-02)
& 19.745 (1.060e-02) & 19.745 (1.060e-02) & 19.745 (1.060e-02) \\
\midrule
Ours (Best RP-GFRFT)
& \textbf{46.721 (2.127e-05)}$^{\mathrm{IY}}$ & \textbf{43.911 (4.064e-05)}$^{\mathrm{IIY}}$ & \textbf{41.392 (7.258e-05)}$^{\mathrm{IIP}}$
& \textbf{42.540 (5.571e-05)}$^{\mathrm{IIP}}$ & \textbf{40.624 (8.662e-05)}$^{\mathrm{IIY}}$ & \textbf{39.619 (1.092e-04)}$^{\mathrm{IIY}}$
& \textbf{42.841 (5.198e-05)}$^{\mathrm{IIP}}$ & \textbf{39.620 (1.091e-04)}$^{\mathrm{IIP}}$ & \textbf{37.491 (1.782e-04)}$^{\mathrm{IIP}}$
& \textbf{42.475 (5.656e-05)}$^{\mathrm{IIY}}$ & \textbf{40.297 (9.338e-05)}$^{\mathrm{IIY}}$ & \textbf{38.454 (1.428e-04)}$^{\mathrm{IIY}}$ \\
\bottomrule
\end{tabular}%
}
\vspace{3pt}

{\footnotesize \textbf{Note:} Entries are PSNR (MSE), and bold entries denote the best result in each column. \textbf{RP-GFRFT code:} $\mathrm{IY}$=I-RP-GFRFT$_{\mathbf{R}_{\text{yaw}}}$, $\mathrm{IIP}$=II-RP-GFRFT$_{\mathbf{R}_{\text{pitch}}}$, and $\mathrm{IIY}$=II-RP-GFRFT$_{\mathbf{R}_{\text{yaw}}}$.}
\end{table}

\subsection{Discussion}
Across the three categories of denoising experiments, namely real-world time series data, natural image data, and 3D point clouds, the proposed RP-GFRFT framework consistently demonstrates robust performance. To balance accuracy and scalability, we adopted differentiated parameter optimization strategies: grid search was used for small-scale time series data to identify stable parameter configurations, while gradient-based optimization was applied to high-dimensional images and 3D point clouds to ensure computational efficiency. This differentiated approach enables RP-GFRFT to achieve superior trade-offs between noise suppression and the preservation of structure or signal features across all tasks. These results also show that RP-GFRFT can be applied consistently to diverse data modalities, ranging from temporal sequences to visual and point-cloud graph signals, highlighting its versatility as a graph signal denoising framework.

\section{Conclusions}
This study proposes a novel RP-GFRFT framework to address the core limitations of existing graph spectral transformation methods, namely, the lack of rotation-based basis control in the GFRFT and the inconsistency in the theoretical degeneracy of the AGFT. To achieve this goal, the framework fundamentally resolves AGFT’s theoretical degeneracy inconsistency by recursively constructing a degeneracy-friendly rotation matrix family. Based on this, two RP-GFRFT variants (I-RP-GFRFT and II-RP-GFRFT) are designed, with strict theoretical proofs completed to validate their core properties. A joint learnable parameterization strategy integrating fractional order $\alpha$ and rotation angle $\theta$ is simultaneously introduced to align the transform parameters with the structural characteristics of diverse graph signals. For the optimization of signals with different scales, the RP-GFRFT adopts a differentiated parameter optimization scheme: a grid search is selected for small-scale temporal signals to ensure parameter reliability, whereas gradient descent is employed for high-dimensional signals to balance efficiency and performance. Multimodal experiments demonstrate the consistent advantages of RP-GFRFT under the tested denoising settings. In future work, we will focus on several directions: extending RP-GFRFT to dynamic graph signal processing, deeply integrating with GNNs to boost spectral feature extraction, optimizing computational efficiency for large-scale scenarios, and expanding to multi-task processing to enhance cross-field practical value.

\bibliographystyle{elsarticle-num}
\bibliography{cas-refs}

\end{document}